\documentclass[11pt]{article}

\usepackage[preprint]{acl}

\usepackage{times}
\usepackage{latexsym}
\usepackage{array}
\usepackage{latexsym}
\usepackage{amsmath}
\usepackage{multirow}
\usepackage{booktabs}
\usepackage{makecell} 
\usepackage[table]{xcolor} 
\usepackage{tabularx}  
\usepackage{tcolorbox}
\usepackage{graphicx}
\tcbuselibrary{breakable, skins, listings}
\usepackage{listings}
\usepackage{longtable}
\usepackage{cuted}   
\usepackage{amssymb}
\usepackage{makecell}
\usepackage{adjustbox}
\newsavebox{\humanPaperQualityBox}

\definecolor{oursblue}{RGB}{232,243,252}
\definecolor{overallorange}{RGB}{255,237,213}

\newcommand{\ourscell}[1]{\cellcolor{oursblue}#1}

\tcbuselibrary{listings,breakable}
\newcolumntype{C}[1]{>{\centering\arraybackslash}p{#1}}
\usepackage[T1]{fontenc}

\usepackage[utf8]{inputenc}

\usepackage{microtype}
\usepackage{hyperref}
\usepackage{inconsolata}

\usepackage{enumitem}
\newtcolorbox{agentprompt}[1]{%
  breakable,
  width=\linewidth,
  colback=white,
  colbacktitle=white,
  colframe=black,
  coltitle=black,
  coltext=black,
  boxrule=0.8pt,
  toprule=0.8pt,
  bottomrule=0.8pt,
  leftrule=0.8pt,
  rightrule=0.8pt,
  titlerule=0.4pt,
  arc=0pt,
  outer arc=0pt,
  title={#1},
  fonttitle=\normalfont\bfseries,
  fontupper=\normalfont,
  boxsep=0pt,
  left=6pt,
  right=6pt,
  top=5pt,
  bottom=5pt,
  toptitle=4pt,
  bottomtitle=4pt,
  before skip=10pt,
  after skip=10pt,
  before upper={\raggedright\setlength{\parindent}{0pt}\setlength{\parskip}{4pt}}
}

\newlist{promptitems}{itemize}{1}
\setlist[promptitems]{
  label=\textbullet,
  leftmargin=1.3em,
  topsep=2pt,
  itemsep=2pt,
  parsep=0pt,
  partopsep=0pt
}

\newlist{promptsteps}{enumerate}{1}
\setlist[promptsteps]{
  label=\arabic*.,
  leftmargin=1.6em,
  topsep=2pt,
  itemsep=2pt,
  parsep=0pt,
  partopsep=0pt
}

\usepackage{booktabs}
\usepackage{listings}
\usepackage{caption}
\title{Discuss More, Comment Less: Boosting Automated Research \\ via Multi-Perspective Collaborative Review}

\author{
Junkai Zhou,
Jiajia Wu,
Zhaoyi Zhang,
Shiting Guan,
Chuanguang Yang
}

\begin{document}
\maketitle
\begin{abstract}
As automated research is increasingly adopted for writing papers, automated review has become a critical link between paper generation and quality control.\ However, existing methods largely rely on single-model feedback or simple review aggregation, failing to explicitly model expert responsibilities or collective deliberation in actual peer reviews, which limits effective cross-examination and reliable decision-making.\ Therefore, we formulate paper review as a computational modeling problem of peer review mechanisms and propose \textbf{RoundtableReview}, a deliberative multi-agent framework for enhancing automated research.\ RoundtableReview structures each paper into eight semantic slots and coordinates six academic and industry reviewers with complementary perspectives, retrieval sources, and specialized responsibilities. Through independent assessment, cross-examination, roundtable deliberation, dynamically weighted voting, and chair synthesis, the framework generates evidence-based and traceable feedback. Experiments on two automated research backbones show that the method improves paper quality in both backbones and outperforms baselines in automatic and human evaluations with fewer comments.
\end{abstract}

\section{Introduction}
Large language models (LLMs) are advancing automated research toward end-to-end scientific workflows spanning literature review, experimental analysis, and paper writing~\citep{lu2024aiscientist,yamada2025aiscientistv2,schmidgall-etal-2025-agent}. Recent work further incorporates peer review into this loop~\citep{liu2025vision,weng2025cycleresearcher}, making automated review central to research quality control. While LLM reviewers can provide useful critiques and well-supported feedback~\citep{liang2024useful, idahl-ahmadi-2025-openreviewer,zhu-etal-2025-deepreview}, reliable final judgments still require cross-validating from multiple perspectives, resolving conflicts, and filtering unsupported comments.

\begin{figure}
    \centering 
    \includegraphics[width=7.6cm]{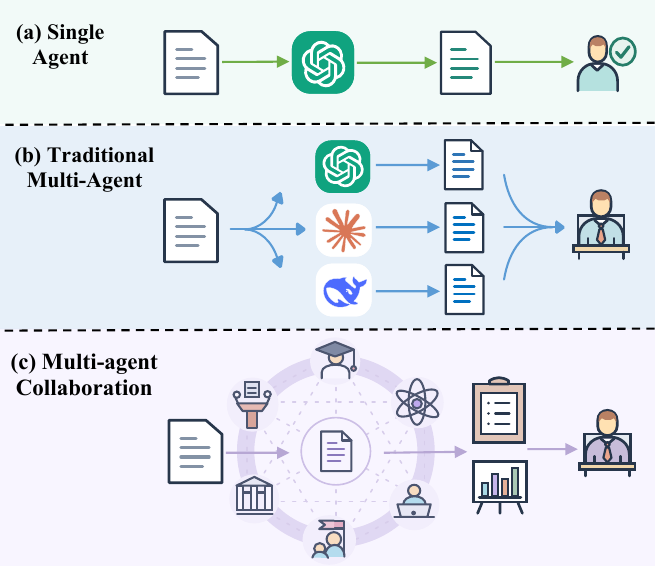}
    \caption{
    When reviewing papers, (a) relies on a single agent with manual verification, (b) aggregates independent reviews with limited interaction, whereas (c) supports specialized multi-agent debate, comment revision and voting to produce a robust and traceable report.
    }
    \label{fig:fig1}
\end{figure}

Existing automated review methods lack modeling of peer review mechanisms in the real world.\ Single-model judgments may miss key claims~\citep{ou-etal-2025-claimcheck,shin-etal-2025-mind}, while multi-agent approaches with role specialization and tools~\citep{darcy2024marg,goyal2026scholarpeer} often rely on simple aggregation, lacking clear responsibility boundaries and overlooking role and expertise differences of experts, thus producing low-quality and redundant feedback, as shown in Figure~\ref{fig:fig1}.\ Therefore, we formulate automated paper review as a computational modeling of peer-review mechanisms, where independent judgments are cross-examined to produce collective decisions with adoption rationales.

Building on this perspective, we propose a deliberative multi-agent framework \textbf{RoundtableReview}, which models automated paper review as a process from evidence-grounding to collective decision-making.\ RoundtableReview structures each paper into eight semantic slots and assigns six academic and industry reviewers with complementary perspectives.\ Each reviewer conducts independent review according to the primary perspective, secondary perspective and explicit responsibility boundaries, with dedicated retrieval sources supporting the primary perspective.\ Reviewers then cross-examine and revise comments through roundtable discussion.\ Surviving comments are selected through voting weighted by role authority and scenario relevance, and a review chairman synthesizes them into the final review.

We then evaluate RoundtableReview on PaperWritingBench~\citep{song2026paperorchestra} with two writing backbones, PaperOrchestra~\citep{song2026paperorchestra} and ARIS~\citep{yang2026aris}, against GLM-5.2~\citep{zeng2026glm} and AI Scientist-v2~\citep{yamada2025aiscientistv2} using longitudinal and horizontal comparisons, automatic evaluators, and blind human evaluation. Despite generating fewer comments than the baselines, RoundtableReview consistently improves both literature review quality and overall paper quality, with notable gains in scientific depth, technical execution, and presentation of evidence. Ablation studies demonstrate that both multi-role discussion and weighted voting contribute to the effectiveness of the framework. The main contributions are as follows:
\begin{itemize}

\item Reconceptualizing automated paper review as a collaborative deliberative process, facilitating traceable cross-examination and collective decision-making 

\item Proposing RoundtableReview that incorporates semantic slots, specialized reviewer roles, retrieval-based perspectives, deliberation, weighted voting, and chair synthesis

\item Evaluating RoundtableReview with two writing backbones and review baselines through automatic and human evaluation, demonstrating improvements in paper quality
\end{itemize}



\section{Related Work}
\label{sec:related-work}
We review prior studies in three research areas closely related to our work: automated research, scientific review, and multi-agent debate.

\paragraph{Automated Research}
Automated research has evolved from task-specific agents to closed-loop workflows covering ideation, experimentation, writing, and review~\citep{lu2024aiscientist,yamada2025aiscientistv2,baek-etal-2025-researchagent,schmidgall-etal-2025-agent}. PaperOrchestra~\citep{song2026paperorchestra}, ARIS~\citep{yang2026aris}, and AutoResearchClaw~\citep{liu2026autoresearchclaw} further advance self-improvement and long-horizon research workflow. Yet weakly validated reviews may misguide revision and limit research quality~\citep{tang2026airesearchagentsnarrow,fei2026agentsfailautoresearchendtoend}. We address the above bottleneck through deliberation and review-guided revision.

\paragraph{Scientific Review}
Scientific review has progressed toward more criteria-based and human-aligned evaluation. LLM reviewers can identify certain issues raised by human~\citep{liang2024useful}, while OpenReviewer and DeepReview improve specificity through specialized training and reasoning~\citep{idahl-ahmadi-2025-openreviewer,zhu-etal-2025-deepreview}. However, alignment gains from guidelines and retrieval vary across models~\citep{li-etal-2026-evaluating-impact,chen-etal-2026-peercheck}, while limited modeling of real-world peer review mechanisms may lead to unsupported or low-quality critiques~\citep{ou-etal-2025-claimcheck,shin-etal-2025-mind,dycke-gurevych-2026-automatic}.

\paragraph{Multi-Agent Debate}
Multi-agent debate improves reasoning through argument exchange, negotiation, and heterogeneous interaction~\citep{du2023debate,chen-etal-2024-reconcile,liang-etal-2024-encouraging}, enabling agents to critically examine competing perspectives, resolve disagreements, and refine their judgments through iterative discussion. However, its effectiveness depends heavily on interaction design and the coordination of diverse agent perspectives~\citep{wang-etal-2024-rethinking-bounds}. ~\citet{kaesberg2025voting} highlight how voting and consensus protocols shape collective reasoning outcomes. In scientific review, MARG and ScholarPeer introduce role specialization and structured interaction~\citep{darcy2024marg,goyal2026scholarpeer}, whereas RoundtableReview explicitly models real-world peer-review mechanisms through evidence-grounded deliberation, systematically tracking how comments are challenged, revised, and incorporated into collective decisions.

\begin{figure*}
    \centering 
    \includegraphics[width=16cm]{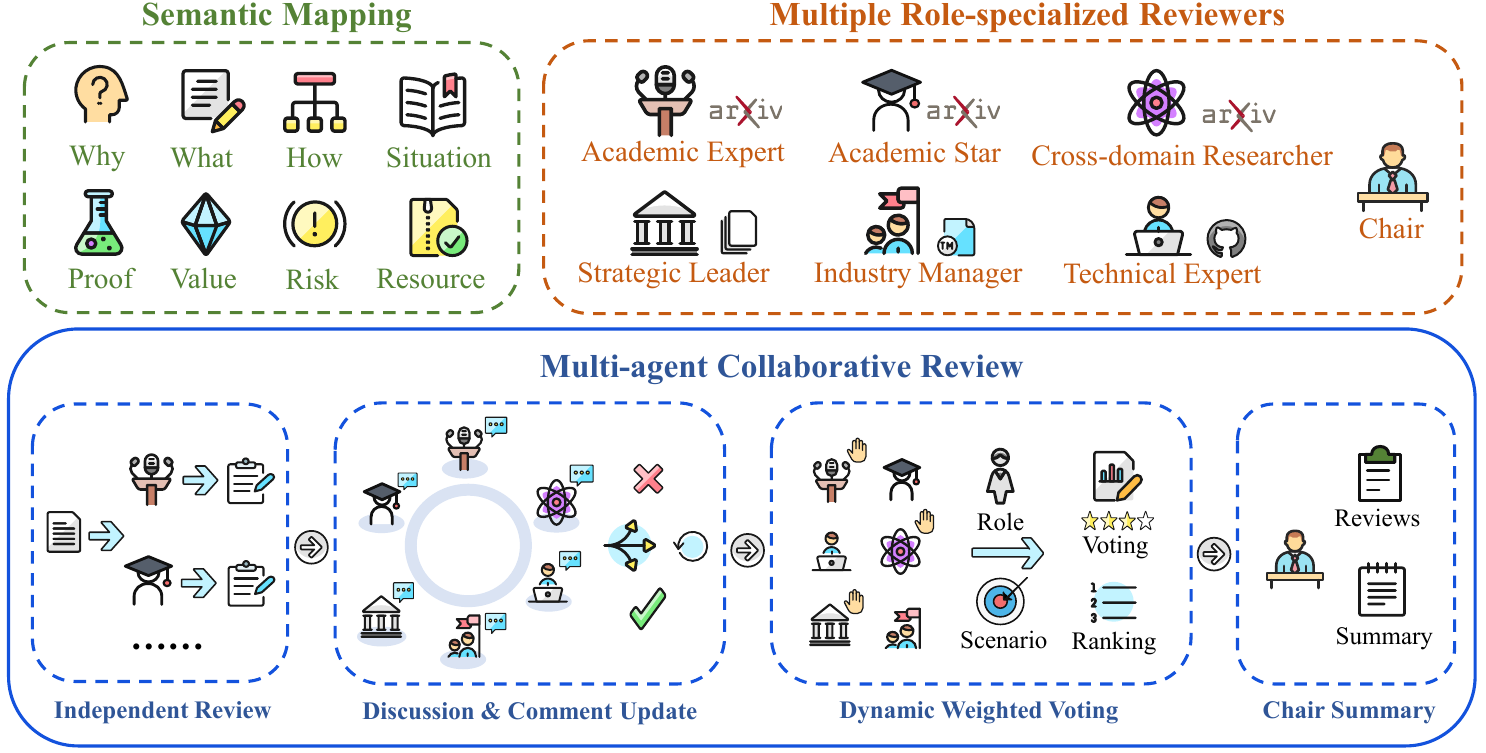}
    \caption{
    The overall framework consists of three parts: (1) paper decomposition into eight semantic slots; (2) six role-specialized reviewer agents and one chair agent; (3) a collaborative multi-agent review process for transforming independent comments into collective reviews through debate, revision, voting, and comment synthesis.}
    \label{fig:fig2}
\end{figure*}

\section{Method}
We propose RoundtableReview, a multi-agent peer review framework designed to enhance automated research. Each paper is decomposed into eight semantic slots for independent review by six role-specialized reviewers with distinct expertise. Their initial reviews are refined through roundtable deliberation, followed by dynamic weighted voting. Review chairman summarizes the accepted comments into a consolidated report. The overall framework is shown in Figure~\ref{fig:fig2}.

\subsection{Task Formulation}
We formalize automated paper review as a deliberative decision-making process. Given a paper $x$, RoundtableReview produces a final review report $y$ and an audit trail $\mathcal{T}$:
\begin{equation}
x\rightarrow \mathcal{Z}\rightarrow \mathcal{O}^{0}
\rightarrow \mathcal{D}\rightarrow \mathcal{O}^{*}
\rightarrow \mathcal{V}\rightarrow (y,\mathcal{T}),
\end{equation}
where $\mathcal{Z}$ denotes semantic slots; $\mathcal{O}^{0}$ and $\mathcal{O}^{*}$ denote opinions before and after deliberation, respectively; $\mathcal{D}$ denotes discussion records; $\mathcal{V}$ denotes the weighted voting results used to select comments; and $\mathcal{T}$ records evidence, discussion, comment evolution, and voting decisions.

\subsection{Semantic Slots}
\label{sec:slots}
To reduce claim-evidence misalignment, we structure content by semantic function instead of section headings. Each paper is mapped into eight slots: background and motivation (\textit{Why}), problem and objective (\textit{What}), related work and research landscape (\textit{Situation}), method and technical approach (\textit{How}), experiments and evidence (\textit{Proof}), value and impact (\textit{Value}), risks and limitations (\textit{Risk}), and resources and reproducibility (\textit{Resource}). For each slot $z_k$, the parser records relevant text spans and locations while retaining full paper access. The process supports role-specific evidence retrieval while mitigating opinion convergence induced by repeated full paper reading.

\subsection{Role-Specialized Reviewers}
\label{sec:roles}
To obtain complementary assessments while controlling for backbone capability, we instantiate six specialized reviewer roles on the same model. Each role adopts one primary perspective supported by retrieval aligned with its responsibilities and two secondary perspectives without external retrieval. Each role also prioritizes semantic slots relevant to its assessment focus. This design combines assessments grounded in external evidence with broader coverage of paper quality. Table~\ref{tab:roles} summarizes the perspectives, priority slots, and retrieval sources assigned to each role.

\begin{table*}[t]
\centering
\small
\renewcommand{\arraystretch}{1.4}
\setlength{\tabcolsep}{3pt}

\resizebox{\textwidth}{!}{
\begin{tabular}{@{}lllll@{}}
\toprule
\textbf{Role}
& \textbf{Primary Perspective}
& \textbf{Secondary Perspectives}
& \textbf{Priority Slots}
& \textbf{Retrieval Source} \\
\midrule

Academic Expert
& research foresight
& contribution \& differentiation
& \textit{Why, What, Situation, How, Value}& arXiv surveys \\

Academic Star
& innovation and depth
& methodological rigor \& soundness
& \textit{What, How, Proof}& arXiv frontier papers \\

Cross-domain Researcher
& cross-domain transfer validity
& completeness \& generalizability
& \textit{What, How, Value}
& out-of-domain arXiv papers \\

Strategic Leader
& strategic value \& goal alignment
& risk controllability \& research capacity
& \textit{Why, What, How, Value, Risk}
& CAPC-CG policy corpus \\

Industry Manager
& compliance with standards
& logical soundness \& clarity
& \textit{What, How, Proof}& EPO OPS \\

Technical Expert
& soundness \& feasibility
& complexity \& resource allocation
& \textit{How, Proof, Resource}& GitHub REST API \\

\midrule

Review Chairman
& \multicolumn{2}{l}{
    Synthesizes approved comments without reviewing or voting.
  }
& All slots
& --- \\

\bottomrule
\end{tabular}
}

\caption{Reviewer roles, perspectives, priority slots, and retrieval sources in RoundtableReview.}
\label{tab:roles}
\end{table*}

The \textbf{academic roles} include an academic expert, an academic star, and a cross-domain researcher, focusing on research foresight, depth of innovation, and cross-domain transferability.\ These roles retrieve public papers through the arXiv API\footnote{\url{https://arxiv.org/}}: survey papers for the academic expert, frontier research excluding surveys for the academic star, and out-of-domain papers for the cross-domain researcher. The \textbf{industry roles} comprise a strategic leader, an industry manager, and a technical expert, focusing on strategic value, standards compliance, and engineering feasibility, respectively. They retrieve sources from CAPC-CG policy corpus~\citep{sun-etal-2026-capc}, the European Patent Office (EPO) OPS API\footnote{\url{https://www.epo.org/en}}, and the GitHub REST API\footnote{\url{https://github.com/}}.

\subsection{Independent Review}
To reduce conformity and premature consensus, all agents first review independently. Each agent follows its role identity, primary and secondary perspectives, priority semantic slots, and predefined instructions, with retrieved evidence added only for the primary perspective, and each agent produces at most three comments. We further introduce a ``honesty protection'' rule: limitations explicitly acknowledged by the authors in \textit{Limitations} or \textit{Future Work} cannot be restated as weaknesses. We represent each initial comment as:
\begin{equation}
o_i^0=(u_i,c_i,a_i),
\end{equation}
where $u_i$, $c_i$, and $a_i$ respectively indicate the issue, rationale, and suggested revision.


\subsection{Roundtable Discussion}
\label{sec}
To improve the quality of initial comments and address role-specific blind spots, we structure roundtable discussion into two stages: scrutiny across roles and comment refinement based on feedback. During cross-role scrutiny, each agent evaluates all comments from the other roles and labels them as \textit{agreement}, \textit{needs-improvement}, or \textit{disagreement}; the latter two require explicit justifications. Comments opposed by at least three of their five peer reviewers are discarded.

During comment refinement, the remaining comments are returned to their proposers, who retain, revise, or withdraw them based on the discussion feedback. Only retained and revised comments proceed to weighted voting session. Further details are provided in Appendix~\ref{sec:app_roundtable}.

\subsection{Dynamic Weighted Voting}
Each agent automatically assigns $+1$ to its prior \textit{agreement} comments. For other comments, it votes based on the revised content, with at most three positive votes in total. All comments not receiving a positive vote from that agent receive $-1$. 

To account for differences in role expertise and scenario relevance, we apply dynamic weighted voting to comments. The vote is weighted by role authority $\alpha_j$ and scenario relevance $\beta_{j,d}$, where $\alpha_j$ captures the relative influence assigned to agent $j$ based on its role expertise, and $\beta_{j,d}$ reflects the alignment between its role and the evaluation priorities of scenario $d$. The weighted vote score is computed as follows:

\begin{equation}
S(o_i^*)=\sum_{j\neq p(i)}\alpha_j\beta_{j,d}v_{ji},
\label{eq:score}
\end{equation}
where $v_{ji}\in\{-1,+1\}$ and $p(i)$ denotes the proposer, comments with $S(o_i^*)\leq 0$ are discarded.



\begin{table*}[t]
\centering
\caption{
Comparison of R1 and R2 rated by three LLM reviewers.
Each cell reports R1 Win / Tie / R2 Win (\%).
Bold marks the highest R2 win rate for each metric within each
backbone and judge group, including ties.
}
\label{tab:r1_r2_results}
\begingroup
\setlength{\tabcolsep}{2pt}
\renewcommand{\arraystretch}{1.35}
\scriptsize
\resizebox{\textwidth}{!}{%
\begin{tabular}{@{}ccc
  *{3}{c}
  >{\columncolor{overallorange}}c
  *{5}{c}
  >{\columncolor{overallorange}[\tabcolsep][0pt]}c@{}}
\toprule
\multirow[c]{3}{*}{Backbone}
& \multirow[c]{3}{*}{Judge}
& \multirow[c]{3}{*}{Reviewer}
& \multicolumn{4}{c}{Literature Review Quality}
& \multicolumn{6}{c}{Paper Quality} \\
\cmidrule(lr){4-7}
\cmidrule(lr){8-13}
& &
& Organization
& Positioning
& Relevance
& Overall Literature
& Scientific
& Technical
& Logical
& Clarity of
& Presentation
& Overall Paper \\
& &
& \& Writing
& \& Novelty
& \& Focus
& Review Quality
& Depth
& Execution
& Flow
& Writing
& of Evidence
& Quality \\
\midrule

\multirow[c]{9}{*}{ARIS}
& \multirow[c]{3}{*}{claude-fable-5}
& GLM-5.2
& 0.0/37.5/62.5
& 2.5/17.5/\textbf{80.0}
& 0.0/52.5/\textbf{47.5}
& 2.5/12.5/\textbf{85.0}
& 0.0/12.5/87.5
& 0.0/27.5/\textbf{72.5}
& 0.0/80.0/20.0
& 7.5/40.0/52.5
& 0.0/12.5/\textbf{87.5}
& 0.0/10.0/90.0 \\

& & AIS-v2
& 0.0/27.5/\textbf{72.5}
& 0.0/35.0/65.0
& 0.0/72.5/27.5
& 0.0/15.0/\textbf{85.0}
& 0.0/10.0/\textbf{90.0}
& 0.0/30.0/70.0
& 0.0/72.5/\textbf{27.5}
& 15.0/62.5/22.5
& 0.0/52.5/47.5
& 0.0/10.0/90.0 \\

& & \ourscell{Ours}
& \ourscell{0.0/90.0/10.0}
& \ourscell{0.0/47.5/52.5}
& \ourscell{2.5/72.5/25.0}
& 2.5/27.5/70.0
& \ourscell{0.0/45.0/55.0}
& \ourscell{0.0/67.5/32.5}
& \ourscell{0.0/77.5/22.5}
& \ourscell{0.0/37.5/\textbf{62.5}}
& \ourscell{0.0/32.5/67.5}
& 0.0/7.5/\textbf{92.5} \\

\cmidrule(lr){2-13}

& \multirow[c]{3}{*}{gpt-5.6-sol}
& GLM-5.2
& 42.5/35.0/\textbf{22.5}
& 2.5/15.0/\textbf{82.5}
& 45.0/42.5/\textbf{12.5}
& 37.5/20.0/\textbf{42.5}
& 0.0/12.5/87.5
& 0.0/25.0/75.0
& 27.5/35.0/37.5
& 50.0/30.0/20.0
& 0.0/27.5/72.5
& 2.5/22.5/75.0 \\

& & AIS-v2
& 67.5/22.5/10.0
& 5.0/22.5/72.5
& 85.0/15.0/0.0
& 65.0/20.0/15.0
& 0.0/27.5/72.5
& 0.0/55.0/45.0
& 37.5/50.0/12.5
& 75.0/17.5/7.5
& 0.0/60.0/40.0
& 27.5/22.5/50.0 \\

& & \ourscell{Ours}
& \ourscell{12.5/75.0/12.5}
& \ourscell{7.5/65.0/27.5}
& \ourscell{5.0/85.0/10.0}
& 7.5/70.0/22.5
& \ourscell{0.0/0.0/\textbf{100.0}}
& \ourscell{0.0/12.5/\textbf{87.5}}
& \ourscell{12.5/37.5/\textbf{50.0}}
& \ourscell{27.5/50.0/\textbf{22.5}}
& \ourscell{0.0/10.0/\textbf{90.0}}
& 0.0/2.5/\textbf{97.5} \\

\cmidrule(lr){2-13}

& \multirow[c]{3}{*}{qwen3.8-max}
& GLM-5.2
& 40.0/35.0/\textbf{25.0}
& 0.0/17.5/\textbf{82.5}
& 32.5/42.5/\textbf{25.0}
& 27.5/25.0/\textbf{47.5}
& 0.0/12.5/87.5
& 0.0/22.5/77.5
& 45.0/37.5/17.5
& 40.0/40.0/20.0
& 0.0/20.0/80.0
& 5.0/15.0/80.0 \\

& & AIS-v2
& 45.0/32.5/22.5
& 5.0/20.0/75.0
& 52.5/30.0/17.5
& 37.5/25.0/37.5
& 2.5/30.0/67.5
& 12.5/37.5/50.0
& 45.0/32.5/22.5
& 72.5/15.0/12.5
& 7.5/40.0/52.5
& 37.5/10.0/52.5 \\

& & \ourscell{Ours}
& \ourscell{15.0/62.5/22.5}
& \ourscell{2.5/62.5/35.0}
& \ourscell{10.0/75.0/15.0}
& 10.0/55.0/35.0
& \ourscell{0.0/5.0/\textbf{95.0}}
& \ourscell{0.0/12.5/\textbf{87.5}}
& \ourscell{7.5/20.0/\textbf{72.5}}
& \ourscell{7.5/27.5/\textbf{65.0}}
& \ourscell{0.0/12.5/\textbf{87.5}}
& 0.0/2.5/\textbf{97.5} \\

\midrule

\multirow[c]{9}{*}{PaperOrchestra}
& \multirow[c]{3}{*}{claude-fable-5}
& GLM-5.2
& 12.5/72.5/\textbf{15.0}
& 5.0/80.0/15.0
& 0.0/90.0/10.0
& 12.5/60.0/27.5
& 10.0/40.0/50.0
& 0.0/50.0/50.0
& 10.0/55.0/35.0
& 7.5/42.5/\textbf{50.0}
& 5.0/57.5/37.5
& 12.5/27.5/60.0 \\

& & AIS-v2
& 12.5/82.5/5.0
& 7.5/80.0/12.5
& 2.5/92.5/5.0
& 10.0/72.5/17.5
& 17.5/45.0/37.5
& 0.0/72.5/27.5
& 10.0/57.5/32.5
& 10.0/65.0/25.0
& 5.0/70.0/25.0
& 12.5/42.5/45.0 \\

& & \ourscell{Ours}
& \ourscell{17.5/72.5/10.0}
& \ourscell{5.0/55.0/\textbf{40.0}}
& \ourscell{15.0/60.0/\textbf{25.0}}
& 17.5/40.0/\textbf{42.5}
& \ourscell{2.5/7.5/\textbf{90.0}}
& \ourscell{2.5/22.5/\textbf{75.0}}
& \ourscell{10.0/40.0/\textbf{50.0}}
& \ourscell{17.5/52.5/30.0}
& \ourscell{10.0/35.0/\textbf{55.0}}
& 5.0/7.5/\textbf{87.5} \\

\cmidrule(lr){2-13}

& \multirow[c]{3}{*}{gpt-5.6-sol}
& GLM-5.2
& 5.0/75.0/\textbf{20.0}
& 15.0/67.5/17.5
& 2.5/65.0/\textbf{32.5}
& 10.0/60.0/30.0
& 10.0/10.0/80.0
& 10.0/17.5/72.5
& 10.0/35.0/\textbf{55.0}
& 10.0/55.0/\textbf{35.0}
& 17.5/42.5/40.0
& 17.5/7.5/75.0 \\

& & AIS-v2
& 7.5/82.5/10.0
& 22.5/65.0/12.5
& 5.0/77.5/17.5
& 12.5/70.0/17.5
& 20.0/22.5/57.5
& 15.0/42.5/42.5
& 20.0/57.5/22.5
& 17.5/75.0/7.5
& 20.0/47.5/32.5
& 30.0/12.5/57.5 \\

& & \ourscell{Ours}
& \ourscell{15.0/82.5/2.5}
& \ourscell{5.0/60.0/\textbf{35.0}}
& \ourscell{2.5/85.0/12.5}
& 10.0/52.5/\textbf{37.5}
& \ourscell{2.5/0.0/\textbf{97.5}}
& \ourscell{0.0/15.0/\textbf{85.0}}
& \ourscell{12.5/35.0/52.5}
& \ourscell{32.5/40.0/27.5}
& \ourscell{2.5/30.0/\textbf{67.5}}
& 2.5/10.0/\textbf{87.5} \\

\cmidrule(lr){2-13}

& \multirow[c]{3}{*}{qwen3.8-max}
& GLM-5.2
& 15.0/52.5/\textbf{32.5}
& 20.0/52.5/27.5
& 7.5/62.5/\textbf{30.0}
& 17.5/50.0/32.5
& 15.0/2.5/82.5
& 7.5/17.5/\textbf{75.0}
& 12.5/27.5/60.0
& 7.5/15.0/\textbf{77.5}
& 10.0/32.5/57.5
& 12.5/0.0/87.5 \\

& & AIS-v2
& 17.5/70.0/12.5
& 25.0/57.5/17.5
& 7.5/80.0/12.5
& 22.5/52.5/25.0
& 30.0/12.5/57.5
& 12.5/47.5/40.0
& 25.0/27.5/47.5
& 25.0/37.5/37.5
& 25.0/25.0/50.0
& 30.0/10.0/60.0 \\

& & \ourscell{Ours}
& \ourscell{22.5/45.0/\textbf{32.5}}
& \ourscell{15.0/32.5/\textbf{52.5}}
& \ourscell{7.5/65.0/27.5}
& 17.5/30.0/\textbf{52.5}
& \ourscell{2.5/2.5/\textbf{95.0}}
& \ourscell{2.5/30.0/67.5}
& \ourscell{5.0/15.0/\textbf{80.0}}
& \ourscell{10.0/25.0/65.0}
& \ourscell{2.5/25.0/\textbf{72.5}}
& 5.0/5.0/\textbf{90.0} \\

\bottomrule
\end{tabular}%
}
\endgroup
\end{table*}

\subsection{Chair Synthesis}
An independent chair synthesizes comments approved through voting into the final report without participating in previous sessions. Using weighted scores, discussion records, and evidence anchors, it merges duplicates and prioritizes the remaining comments. The report contains an overall assessment and 3-6 recommendations. Complete prompt specifications for all agents and all stages of the framework are provided in Appendix~\ref{app:prompts}.

\section{Experiments}
\label{sec:experiments}

\subsection{Experimental Settings}
\paragraph{Dataset}
PaperWritingBench~\citep{song2026paperorchestra} is adopted, with 20 papers randomly sampled from each of CVPR 2025 and ICLR 2025, totaling 40 papers. Each sample provides a research idea, experiment logs, conference template, submission guidelines, and visual assets for paper generation.

\paragraph{Implementation Details}
RoundtableReview uses GLM-5.2 via API with temperature = 1.0 and top\_p = 0.95. We set max\_tokens to 120,000 and limit the context window to 1,500,000 characters. The authority weights $\alpha_j$ for the technical expert, strategic leader, industry manager, cross-domain researcher, academic expert, and academic star roles are set to 1.1, 1.5, 1.0, 1.1, 1.5, and 1.0, respectively. The relevance weights $\beta_{j,\text{paper}}$ are set to 1.2 for academic roles and 0.9 for industry roles. These weights were selected through parameter tuning. The total cost across all experiments and evaluations is approximately \$2000 USD. 

For the retrieval sources supporting the primary perspective of each agent, the academic expert retrieves 5 arXiv surveys, while the academic star retrieves 8 arXiv papers and retains up to 5 non-survey research papers.\ The cross-domain researcher retrieves 5 arXiv papers. For these agents, HTML full texts are truncated to 40,000 characters, with abstracts used only when full text retrieval fails. The strategic leader retrieves 10 skeleton entries from the CAPC-CG policy corpus. The industry manager retrieves 5 European and international patents via the EPO OPS API, each truncated to the 3,000 characters. The technical expert retrieves 10 GitHub repositories ranked by stars, using their descriptions and metadata.


\paragraph{Backbones, Review Baselines, and Ablations}
We use PaperOrchestra~\citep{song2026paperorchestra}, reproduced from its public code with Gemini 3.1 Pro, and ARIS~\citep{yang2026aris}, using its paper-writing skill with Claude Sonnet 4.6. For both backbone agents, we retain the original parameter settings. They generate R1 from identical inputs, then revise it into R2 based on review feedback, and all review methods share the same R1 input.

We compare against two review baselines: GLM-5.2~\citep{zeng2026glm} and the review component of AI Scientist-v2 (AIS-v2)~\citep{yamada2025aiscientistv2}. In addition, we evaluate two ablated variants of RoundtableReview: parallel review omits deliberation and retains only independent reviews ($w/o$ deliberation), while equal voting sets $\alpha_j=\beta_{j,d}=1$, removing both role and relevance weighting ($w/o$ weighting).

\subsection{Evaluation Method and Metrics}
We evaluate literature review quality and overall paper quality through automatic and human evaluation, and automatic evaluation uses both LLMs and reviewer agents, following \citet{song2026paperorchestra}.

\begin{table*}[t]
\centering
\caption{
Comparison of R2 manuscripts revised by different methods and rated by three LLM reviewers.
Each cell reports Baseline Win / Tie / Ours Win (\%).
}
\label{tab:reviewer_comparison}
\begingroup
\setlength{\tabcolsep}{2pt}
\renewcommand{\arraystretch}{1.35}
\scriptsize
\resizebox{\textwidth}{!}{%
\begin{tabular}{@{}ccc
  *{3}{c}
  >{\columncolor{overallorange}}c
  *{5}{c}
  >{\columncolor{overallorange}[\tabcolsep][0pt]}c@{}}
\toprule
\multirow[c]{3}{*}{Backbone}
& \multirow[c]{3}{*}{Judge}
& \multirow[c]{3}{*}{Comparison}
& \multicolumn{4}{c}{Literature Review Quality}
& \multicolumn{6}{c}{Paper Quality} \\
\cmidrule(lr){4-7}
\cmidrule(lr){8-13}
& &
& Organization
& Positioning
& Relevance
& Overall Literature
& Scientific
& Technical
& Logical
& Clarity of
& Presentation
& Overall Paper \\
& &
& \& Writing
& \& Novelty
& \& Focus
& Review Quality
& Depth
& Execution
& Flow
& Writing
& of Evidence
& Quality \\
\midrule

\multirow[c]{6}{*}{ARIS}
& \multirow[c]{2}{*}{claude-fable-5}
& GLM-5.2 vs. ours
& 20.0/37.5/\textbf{42.5}
& \textbf{47.5}/20.0/32.5
& 15.0/50.0/\textbf{35.0}
& 32.5/25.0/\textbf{42.5}
& \textbf{47.5}/27.5/25.0
& \textbf{50.0}/22.5/27.5
& 7.5/45.0/\textbf{47.5}
& 7.5/35.0/\textbf{57.5}
& \textbf{50.0}/25.0/25.0
& \textbf{42.5}/15.0/\textbf{42.5} \\

& & AIS-v2 vs. ours
& 12.5/30.0/\textbf{57.5}
& \textbf{42.5}/25.0/32.5
& 12.5/27.5/\textbf{60.0}
& 25.0/17.5/\textbf{57.5}
& 12.5/20.0/\textbf{67.5}
& 12.5/25.0/\textbf{62.5}
& 5.0/30.0/\textbf{65.0}
& 2.5/20.0/\textbf{77.5}
& 15.0/22.5/\textbf{62.5}
& 15.0/7.5/\textbf{77.5} \\

\cmidrule(lr){2-13}

& \multirow[c]{2}{*}{gpt-5.6-sol}
& GLM-5.2 vs. ours
& 20.0/22.5/\textbf{57.5}
& \textbf{70.0}/17.5/12.5
& 27.5/35.0/\textbf{37.5}
& \textbf{40.0}/22.5/37.5
& \textbf{70.0}/12.5/17.5
& \textbf{60.0}/15.0/25.0
& 22.5/10.0/\textbf{67.5}
& 17.5/17.5/\textbf{65.0}
& \textbf{62.5}/20.0/17.5
& \textbf{55.0}/15.0/30.0 \\

& & AIS-v2 vs. ours
& 5.0/25.0/\textbf{70.0}
& \textbf{67.5}/20.0/12.5
& 10.0/20.0/\textbf{70.0}
& 17.5/20.0/\textbf{62.5}
& 35.0/17.5/\textbf{47.5}
& 25.0/20.0/\textbf{55.0}
& 7.5/22.5/\textbf{70.0}
& 2.5/12.5/\textbf{85.0}
& 22.5/27.5/\textbf{50.0}
& 20.0/7.5/\textbf{72.5} \\

\cmidrule(lr){2-13}

& \multirow[c]{2}{*}{qwen3.8-max}
& GLM-5.2 vs. ours
& \textbf{25.0}/62.5/12.5
& \textbf{37.5}/55.0/7.5
& \textbf{27.5}/60.0/12.5
& \textbf{40.0}/47.5/12.5
& \textbf{52.5}/40.0/7.5
& \textbf{35.0}/55.0/10.0
& \textbf{22.5}/67.5/10.0
& \textbf{15.0}/70.0/\textbf{15.0}
& \textbf{40.0}/45.0/15.0
& \textbf{50.0}/32.5/17.5 \\

& & AIS-v2 vs. ours
& \textbf{20.0}/70.0/10.0
& \textbf{27.5}/57.5/15.0
& 7.5/70.0/\textbf{22.5}
& \textbf{30.0}/50.0/20.0
& \textbf{17.5}/70.0/12.5
& 15.0/67.5/\textbf{17.5}
& \textbf{15.0}/72.5/12.5
& 12.5/67.5/\textbf{20.0}
& 10.0/72.5/\textbf{17.5}
& 20.0/52.5/\textbf{27.5} \\

\midrule

\multirow[c]{6}{*}{PaperOrchestra}
& \multirow[c]{2}{*}{claude-fable-5}
& GLM-5.2 vs. ours
& \textbf{30.0}/60.0/10.0
& 10.0/57.5/\textbf{32.5}
& \textbf{15.0}/77.5/7.5
& \textbf{35.0}/35.0/30.0
& 15.0/55.0/\textbf{30.0}
& \textbf{25.0}/57.5/17.5
& \textbf{22.5}/70.0/7.5
& \textbf{30.0}/57.5/12.5
& \textbf{20.0}/60.0/\textbf{20.0}
& \textbf{42.5}/20.0/37.5 \\

& & AIS-v2 vs. ours
& 15.0/67.5/\textbf{17.5}
& 12.5/52.5/\textbf{35.0}
& \textbf{10.0}/80.0/\textbf{10.0}
& 27.5/37.5/\textbf{35.0}
& 5.0/20.0/\textbf{75.0}
& 2.5/52.5/\textbf{45.0}
& 12.5/67.5/\textbf{20.0}
& 20.0/52.5/\textbf{27.5}
& 10.0/52.5/\textbf{37.5}
& 12.5/17.5/\textbf{70.0} \\

\cmidrule(lr){2-13}

& \multirow[c]{2}{*}{gpt-5.6-sol}
& GLM-5.2 vs. ours
& \textbf{27.5}/62.5/10.0
& 15.0/42.5/\textbf{42.5}
& \textbf{25.0}/65.0/10.0
& 25.0/47.5/\textbf{27.5}
& 22.5/20.0/\textbf{57.5}
& \textbf{47.5}/22.5/30.0
& \textbf{22.5}/57.5/20.0
& \textbf{32.5}/50.0/17.5
& 27.5/32.5/\textbf{40.0}
& 40.0/10.0/\textbf{50.0} \\

& & AIS-v2 vs. ours
& 15.0/67.5/\textbf{17.5}
& 10.0/42.5/\textbf{47.5}
& \textbf{17.5}/65.0/\textbf{17.5}
& 17.5/45.0/\textbf{37.5}
& 5.0/5.0/\textbf{90.0}
& 12.5/32.5/\textbf{55.0}
& 10.0/52.5/\textbf{37.5}
& 17.5/60.0/\textbf{22.5}
& 17.5/25.0/\textbf{57.5}
& 15.0/7.5/\textbf{77.5} \\

\cmidrule(lr){2-13}

& \multirow[c]{2}{*}{qwen3.8-max}
& GLM-5.2 vs. ours
& \textbf{30.0}/45.0/25.0
& 15.0/57.5/\textbf{27.5}
& \textbf{15.0}/72.5/12.5
& 30.0/37.5/\textbf{32.5}
& 7.5/37.5/\textbf{55.0}
& 10.0/60.0/\textbf{30.0}
& 2.5/75.0/\textbf{22.5}
& 17.5/55.0/\textbf{27.5}
& 5.0/77.5/\textbf{17.5}
& 17.5/27.5/\textbf{55.0} \\

& & AIS-v2 vs. ours
& 17.5/60.0/\textbf{22.5}
& 17.5/57.5/\textbf{25.0}
& 7.5/70.0/\textbf{22.5}
& 25.0/45.0/\textbf{30.0}
& 2.5/15.0/\textbf{82.5}
& 2.5/40.0/\textbf{57.5}
& 0.0/62.5/\textbf{37.5}
& 10.0/45.0/\textbf{45.0}
& 0.0/52.5/\textbf{47.5}
& 10.0/10.0/\textbf{80.0} \\

\bottomrule
\end{tabular}%
}
\endgroup
\end{table*}

\paragraph{Evaluation Metrics}
Our evaluation places greater emphasis on overall paper quality, with literature review quality serving as a complementary measure. Overall paper quality covers five dimensions: scientific depth, technical execution, logical flow, clarity of writing, and presentation of evidence. Literature review quality covers three dimensions: organization and writing, positioning and novelty, and relevance and focus. More details are provided in Appendix~\ref{sec:app_dimensions}. 

\paragraph{Bidirectional Side-by-Side Comparison}
To mitigate LLM positional bias, each pair of papers is evaluated in both AB/BA orders, with the labeled as a win (two wins, or one win and one tie), a tie (one win and one loss, or two ties), or a loss. Votes are aggregated across the three literature review quality and five paper quality dimensions, and the overall result is their average. Evaluation uses three LLMs (claude-fable-5, gpt-5.6-sol, and qwen3.8-max) and two reviewer agents (ScholarPeer~\citep{goyal2026scholarpeer} and AutoResearchClaw~\citep{liu2026autoresearchclaw}) under identical settings. We use two complementary evaluation settings: longitudinal evaluation compares R1 and R2 from the same review pipeline to measure revision gains, while horizontal evaluation compares R2 manuscripts produced using different review methods to assess their relative effectiveness.

\paragraph{Human Evaluation}
We randomly sample six papers from ICLR and CVPR. One PhD researcher and two graduate students voluntarily conduct anonymous side-by-side comparisons of the revised R2 papers produced by our method and baselines, following predefined guidelines and using the same metrics as automatic evaluation.






\subsection{Results}
We investigate four key research questions.

\begin{enumerate}
    \item[\textbf{Q1:}] Does RoundtableReview consistently improve paper quality across different writing backbones and evaluators?
    \item[\textbf{Q2:}] Does RoundtableReview outperform review baselines in improving paper quality?
    \item[\textbf{Q3:}] How does review comment quantity affect revision effectiveness?
    \item[\textbf{Q4:}] How do multi-role discussion and weighted voting contribute to overall performance?
\end{enumerate}

\paragraph{Longitudinal Evaluation Results (Q1)}
As shown in Table~\ref{tab:r1_r2_results}, RoundtableReview improves paper quality from R1 to R2 across two writing backbones and three LLM judges. For overall paper quality, RoundtableReview achieves the highest R2 win rate in all settings, outperforming both GLM-5.2 and AI Scientist-v2, with win rates of 92.5\%-97.5\% for ARIS and 87.5\%-90.0\% for PaperOrchestra. The gains are particularly pronounced in scientific depth, technical execution, and presentation of evidence. For overall literature review quality, RoundtableReview achieves the highest performance across all three judges under PaperOrchestra, while showing weaker results under ARIS, likely reflecting differences in how the writing backbones utilize review comments.








\begin{figure*}
    \centering 
    \includegraphics[width=15.5cm]{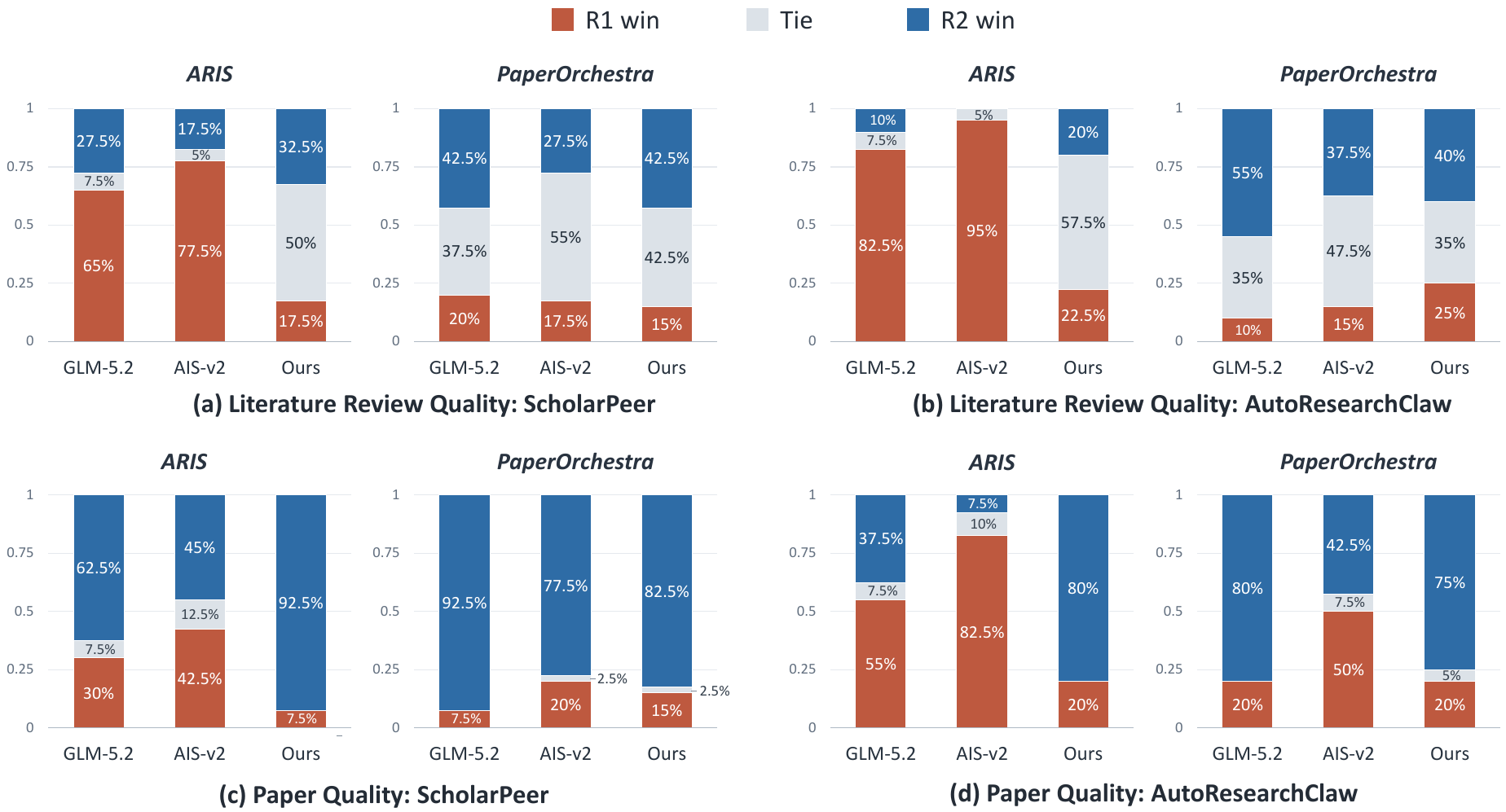}
    \caption{
    Longitudinal evaluation results comparing Round 1 (R1) and Round 2 (R2) on literature review and paper quality: (a) Literature review quality evaluated by ScholarPeer; (b) literature review quality evaluated by AutoResearchClaw; (c) paper quality evaluated by ScholarPeer; (d) paper quality evaluated by AutoResearchClaw.}
    \label{fig:fig3}
\end{figure*}

Agent-based evaluation with ScholarPeer and AutoResearchClaw further supports the effectiveness of RoundtableReview, as shown in Figure~\ref{fig:fig3}. For overall paper quality, RoundtableReview achieves R2 win rates of 75.0\%-92.5\% across the two writing backbones, outperforming AI Scientist-v2 in all settings and showing more consistent gains than GLM-5.2. For overall literature review quality, RoundtableReview also generally outperforms AI Scientist-v2 and exceeds GLM-5.2 in most settings. Detailed scores for each subdimension of the agent-based longitudinal evaluation are reported in Appendix~\ref{sec:longitudinal-dimensions}.

\paragraph{Horizontal Evaluation Results (Q2)}
As demonstrated by the comparasion results presented in Table~\ref{tab:reviewer_comparison}, the advantages of
RoundtableReview are primarily reflected in overall paper quality.
RoundtableReview outperforms AI Scientist-v2 across all six
backbone-judge combinations, with notable gains in scientific depth,
technical execution, and presentation of evidence. Compared with
GLM-5.2, RoundtableReview performs better with PaperOrchestra but worse
with ARIS. Nevertheless, GLM-5.2 yields smaller R1 to R2 gains in the
longitudinal evaluation, indicating that it is less effective than
RoundtableReview in improving the original paper.

\begin{figure*}
    \centering 
    \includegraphics[width=15.5cm]{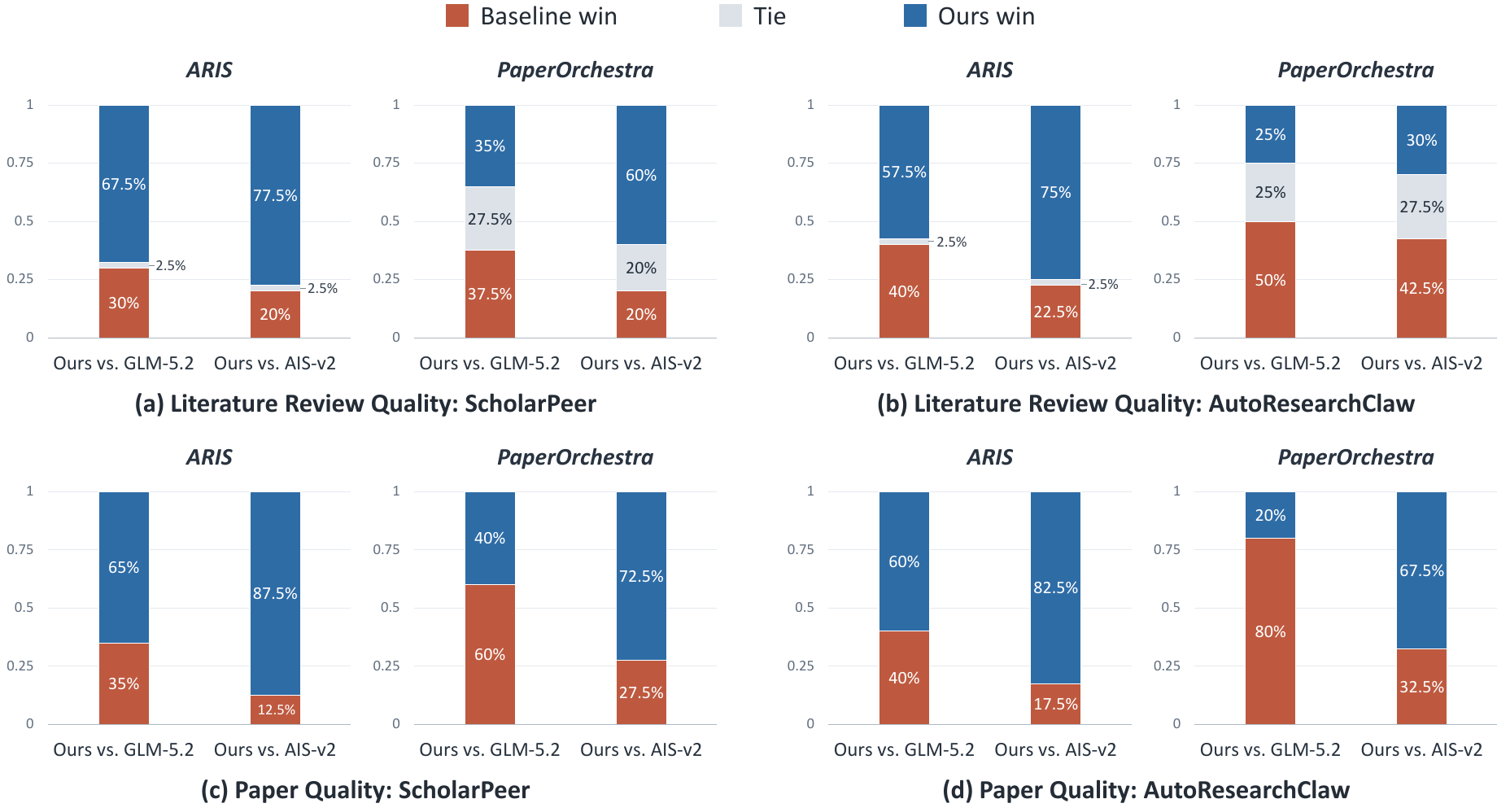}
    \caption{
    Horizontal evaluation results comparing our method with GLM-5.2 and AIS-v2 in R2 using ARIS and PaperOrchestra: (a) literature review quality evaluated by ScholarPeer; (b) literature review quality evaluated by AutoResearchClaw; (c) paper quality evaluated by ScholarPeer; (d) paper quality evaluated by AutoResearchClaw.}
    \label{fig:fig4}
\end{figure*}

Agent-based horizontal evaluation with ScholarPeer and AutoResearchClaw further confirms the advantage of RoundtableReview in overall paper quality, as shown in Figure~\ref{fig:fig4}. RoundtableReview outperforms AI Scientist-v2 across all settings, while it performs better under ARIS and remains more closely matched with GLM-5.2 under PaperOrchestra. For overall literature review quality, RoundtableReview shows stronger performance under ARIS and variation under PaperOrchestra, possibly reflecting greater evaluator sensitivity in literature review assessment. Detailed scores for subdimensions of the agent-based horizontal evaluation are reported in Appendix~\ref{sec:app-horizontal-dimensions}.

\begin{table}[t]
\centering
\caption{
Human evaluation results of R2 manuscripts. Each cell reports Baseline Win / Tie / Ours Win (\%).
}
\label{tab:human_full}
\begingroup
\scriptsize
\setlength{\tabcolsep}{2pt}
\renewcommand{\arraystretch}{1.3}
\renewcommand{\theadfont}{\normalfont}

\sbox{\humanPaperQualityBox}{%
\begin{tabular}{@{}c*{5}{c}
  >{\columncolor{overallorange}[\tabcolsep][0pt]}c@{}}
\toprule
\thead[c]{Comparison}
& \thead[c]{Scientific\\Depth}
& \thead[c]{Technical\\Execution}
& \thead[c]{Logical\\Flow}
& \thead[c]{Clarity of\\Writing}
& \thead[c]{Presentation\\of Evidence}
& \thead[c]{Overall Paper\\Quality} \\
\midrule

GLM-5.2 vs. Ours
& 22.2/44.4/\textbf{33.3}
& 27.8/33.3/\textbf{38.9}
& 0.0/16.7/\textbf{83.3}
& 0.0/11.1/\textbf{88.9}
& 22.2/27.8/\textbf{50.0}
& 11.1/11.1/\textbf{77.8} \\
AIS-v2 vs. Ours
& 16.7/44.4/\textbf{38.9}
& 0.0/27.8/\textbf{72.2}
& 11.1/16.7/\textbf{72.2}
& 5.6/33.3/\textbf{61.1}
& 5.6/33.3/\textbf{61.1}
& 0.0/11.1/\textbf{88.9} \\

\cmidrule(lr){1-7}

GLM-5.2 vs. Ours
& 11.1/27.8/\textbf{61.1}
& 5.6/38.9/\textbf{55.6}
& 0.0/55.6/\textbf{44.4}
& 16.7/27.8/\textbf{55.6}
& 11.1/33.3/\textbf{55.6}
& 0.0/22.2/\textbf{77.8} \\
AIS-v2 vs. Ours
& 16.7/22.2/\textbf{61.1}
& 5.6/22.2/\textbf{72.2}
& 11.1/50.0/\textbf{38.9}
& 11.1/27.8/\textbf{61.1}
& 11.1/33.3/\textbf{55.6}
& 16.7/11.1/\textbf{72.2} \\

\bottomrule
\end{tabular}%
}

\resizebox{\columnwidth}{!}{%
\begin{minipage}{\wd\humanPaperQualityBox}
\centering
\setlength{\parskip}{0pt}

\begin{tabular*}{\linewidth}
{@{\extracolsep{\fill}}cc*{3}{c}
  >{\columncolor{overallorange}[\tabcolsep][0pt]}c@{}}
\toprule
\thead[c]{Writing\\Backbone}
& \thead[c]{Comparison}
& \thead[c]{Organization\\\& Writing}
& \thead[c]{Positioning\\\& Novelty}
& \thead[c]{Relevance\\\& Focus}
& \thead[c]{Overall Literature\\Review Quality} \\
\midrule

\multirowcell{2}{ARIS}
& GLM-5.2 vs. Ours
& 22.2/22.2/\textbf{55.6}
& 33.3/27.8/\textbf{38.9}
& 11.1/33.3/\textbf{55.6}
& 22.2/11.1/\textbf{66.7} \\
& AIS-v2 vs. Ours
& 16.7/27.8/\textbf{55.6}
& 27.8/38.9/\textbf{33.3}
& 5.6/33.3/\textbf{61.1}
& 22.2/16.7/\textbf{61.1} \\

\cmidrule(lr){2-6}

\multirowcell{2}{Paper-\\Orchestra}
& GLM-5.2 vs. Ours
& 11.1/44.4/\textbf{44.4}
& 0.0/50.0/\textbf{50.0}
& 11.1/44.4/\textbf{44.4}
& 5.6/33.3/\textbf{61.1} \\
& AIS-v2 vs. Ours
& 5.6/38.9/\textbf{55.6}
& 11.1/33.3/\textbf{55.6}
& 5.6/44.4/\textbf{50.0}
& 0.0/38.9/\textbf{61.1} \\

\bottomrule
\end{tabular*}

\par\vspace{2pt}

\usebox{\humanPaperQualityBox}

\end{minipage}%
}

\endgroup
\end{table}


\paragraph{Human Evaluation Results}
Human evaluation further validates the effectiveness of RoundtableReview, as shown in Table~\ref{tab:human_full}. Across two writing backbones and two baselines, RoundtableReview consistently outperforms the baselines in both overall assessments and all evaluation dimensions, demonstrating robust improvements across different experimental settings. These gains span scientific depth, technical execution, logical flow, writing clarity, and evidence presentation, indicating broad improvements in manuscript quality.\ Inter-annotator agreement yields a Fleiss' kappa~\citep{fleiss1971measuring} of 0.087. Fleiss' kappa and Cohen's kappa~\citep{cohen1960coefficient} statistics for human-LLM and human-agent agreement are reported in Appendix~\ref{sec:app_Results of Human Evaluation}.

%

%

\paragraph{Further Analysis of RoundtableReview (Q3)}
As shown in Table~\ref{tab:revision_suggestion_count}, RoundtableReview achieves stronger and more stable revision performance with only 4.2-4.3 suggestions per paper on average, compared with 12.7-17.6 for the baselines. This highlights the comment efficiency of RoundtableReview: multi-role discussion and collective selection concentrate revision guidance into a smaller set of more effective suggestions rather than relying on comment volume. In a few settings, the stronger performance of the baselines may partly stem from retaining a larger number of revision suggestions. 

\begin{table}[t]
\centering
\caption{
Number of review comments generated by each review method. Total and Average denote total comment counts and averages per paper, respectively.}
\label{tab:revision_suggestion_count}
\begingroup
\setlength{\tabcolsep}{11pt}
\renewcommand{\arraystretch}{1}
\scriptsize
\begin{tabular}{@{}cccc@{}}
\toprule
Backbone & Method & Total & Average \\
\midrule

\multirow[c]{5}{*}{PaperOrchestra}
& Ours               & 169 & 4.2 \\
& AIS-v2              & 509& 12.7\\
& GLM-5.2             & 606& 15.2\\
& $w/o$ deliberation & 178 & 4.5 \\
& $w/o$ weighting    & 146 & 3.6 \\

\midrule

\multirow[c]{5}{*}{ARIS}
& Ours               & 173 & 4.3 \\
& AIS-v2              & 507& 12.7\\
& GLM-5.2             & 703& 17.6\\
& $w/o$ deliberation & 175 & 4.4 \\
& $w/o$ weighting    & 229 & 5.7 \\

\bottomrule
\end{tabular}
\endgroup
\end{table}

\paragraph{Ablation Study (Q4)}
The longitudinal evaluation results in Table~\ref{tab:ablation_r1_r2} of Appendix~\ref{sec:app_Results of Ablation Study} show that removing either the weighting mechanism or multi-role discussion generally degrades overall performance. Removing the weighting mechanism causes a modest drop, while removing multi-role discussion exhibits stronger backbone dependence, with larger declines under PaperOrchestra, particularly in overall paper quality and technical dimensions. 
The horizontal evaluation results in Table~\ref{tab:ablation_r2_r2} of Appendix~\ref{sec:app_Results of Ablation Study} further demonstrate consistent benefits from the weighting mechanism, while the effect of multi-role discussion varies by backbone. Case studies comparing excerpts from our method, AI-scientist-v2, and GLM-5.2 reviews are presented in Table~\ref{tab:semantic-noise-case-study} of Appendix~\ref{sec:case-study}.





\section{Conclusion}
\label{sec:conclusion}
We propose RoundtableReview, a deliberative multi-agent framework that models real-world peer review mechanisms for evidence-based and traceable automated review. Our framework structures papers into semantic slots and coordinates specialized reviewers through independent review, cross-examination, roundtable deliberation, dynamically weighted voting, and chair synthesis. Across two writing backbones, both automatic and human evaluations demonstrate consistent improvements in paper quality. Notably, RoundtableReview enhances review effectiveness with fewer comments, highlighting its value in delivering more focused and impactful feedback.

\section*{Limitations}
Our framework remains an approximation of real-world peer review to a certain extent. Prompt-based role specialization, single-round deliberation, and predefined voting weights simplify the diversity of reviewer expertise and the dynamics of multi-round discussion. Additionally, although our evaluation combines automatic and human evaluation across multiple settings, the number of human annotators remains limited. Future work may explore richer reviewer and larger-scale expert evaluation in broader research domains. Although evaluated in paper review, RoundtableReview may also be adapted to project reviews, proposal evaluations, and related applications.

\section*{Ethical Considerations}
This study relies on existing LLMs for paper review, discussion, and evaluation, therefore inherits common risks of LLM-based systems, including inaccurate, unsupported, or inappropriate comment. RoundtableReview also relies on external retrieval sources, including public papers, patents, and policy documents. Although these sources do not contain offensive content, the retrieved information may contain errors or biases. All models, datasets, and external resources should be used in accordance with their applicable licenses, terms of use, and data policies.

Role specialization, retrieval, and weighted voting may introduce bias related to discipline, language, expertise, or research conventions. To improve transparency, RoundtableReview preserves evidence sources, discussion records, comment revisions, and voting results, allowing users to inspect how recommendations are formed. Broader evaluation across research domains, languages, and reviewer populations remains necessary before deployment in real world.

RoundtableReview is intended to support manuscript improvement rather than replace human judgment in paper acceptance, academic assessment, or other high-impact decisions. Automated reviews should therefore be interpreted as decision-support signals rather than authoritative judgments, with meaningful human oversight retained in consequential settings. Finally, although this study does not train new foundation models, multi-agent reviewing, retrieval, and repeated evaluation incur additional computational and environmental costs.

\bibliography{custom}

\clearpage

\appendix

\begin{center}
    {\large\bfseries Appendix Contents}
\end{center}

\vspace{0.5em}

\noindent
\textbf{A}\quad
\hyperref[sec:app_roundtable]{More Details of Roundtable Discussion}

\vspace{0.25em}

\noindent
\textbf{B}\quad
\hyperref[app:prompts]{RoundtableReview: Detailed Prompt}

\vspace{0.25em}

\noindent
\textbf{C}\quad
\hyperref[sec:app_dimensions]{More Details of Evaluation Metrics}

\vspace{0.25em}

\noindent
\textbf{D}\quad
\hyperref[sec:longitudinal-dimensions]{More Longitudinal Evaluation Results}

\vspace{0.25em}

\noindent
\textbf{E}\quad
\hyperref[sec:app-horizontal-dimensions]{More Horizontal Evaluation Results}

\vspace{0.25em}

\noindent
\textbf{F}\quad
\hyperref[sec:app_Results of Human Evaluation]{Human Assessment Consistency}



\vspace{0.25em}

\noindent
\textbf{G}\quad
\hyperref[sec:app_Results of Ablation Study]{Results of Ablation Study}

\vspace{0.25em}

\noindent
\textbf{H}\quad
\hyperref[sec:case-study]{Case Study}

\section{More Details of Roundtable Discussion}
\label{sec:app_roundtable}
Each role may agree with at most five comments and must disagree with at least three, with the rest labeled needs-improvement. Excess agreement judgments are converted to needs-improvement, and insufficient disagreement judgments are supplemented from needs-improvement.

\section{RoundtableReview: Detailed Prompt}
\label{app:prompts}

The review pipeline consists of six parallel reviewer agents and one chairman agent,
operating across four stages: (1)~independent review, (2)~structured peer discussion,
(3)~final report aggregation by the chairman.
Stages~2 and~3 use identical prompt templates across all six reviewer agents; only
the agent name and \texttt{role\_description} field are substituted per role.
Section~\ref{sec:a1} provides an agent overview; Sections~\ref{sec:a2}-\ref{sec:a5}
give the complete prompts for each stage.
\subsection{Agent Overview}
\label{sec:a1}
Table~\ref{tab:agents} summarizes the seven agents and their respective roles within RoundtableReview, highlighting their distinct responsibilities and complementary functions throughout the collaborative review process.

\begin{table*}[t]
\centering
\caption{
Roles, evaluation perspectives, and retrieval sources of the seven agents.
}
\label{tab:agents}
\begingroup
\small
\setlength{\tabcolsep}{4pt}
\renewcommand{\arraystretch}{1.2}

\begin{tabularx}{\textwidth}{@{}
  >{\raggedright\arraybackslash}p{0.17\textwidth}
  >{\raggedright\arraybackslash}p{0.16\textwidth}
  >{\raggedright\arraybackslash}p{0.20\textwidth}
  >{\raggedright\arraybackslash}X
  >{\raggedright\arraybackslash}p{0.12\textwidth}
@{}}
\toprule
\textbf{Agent Name}
& \textbf{ID}
& \textbf{Primary Perspective}
& \textbf{Secondary Perspectives}
& \textbf{Retrieval} \\
\midrule

Senior Academic Expert
& \texttt{academic\_expert}
& Academic foresight
& Degree of contribution; comparison with existing work
& arXiv surveys \\

\addlinespace[3pt]

Rising Academic Star
& \texttt{academic\_star}
& Depth of academic innovation
& Methodological rigor; experimental soundness
& arXiv research \\

\addlinespace[3pt]

Cross-Domain Researcher
& \texttt{cross\_domain}
& Cross-domain knowledge and insights
& Methodological completeness; generalizability
& arXiv (adjacent) \\

\addlinespace[3pt]

Strategic Leader
& \texttt{strategic}
& Strategic value and macro-goal alignment
& Risk controllability; positioning correctness
& CAPC-CG \\

\addlinespace[3pt]

Industry Manager
& \texttt{managerial}
& Industry standards compliance
& Logical soundness; goal decomposition; clarity
& EPO OPS \\

\addlinespace[3pt]

Technical Expert
& \texttt{tech\_expert}
& Technical soundness and feasibility
& Engineering complexity; resource reasonableness
& GitHub \\

\midrule

Review Chairman
& \texttt{chairman}
& \multicolumn{3}{
    >{\raggedright\arraybackslash}
    p{\dimexpr0.67\textwidth-4\tabcolsep\relax}@{}
  }{
    Aggregates all reviewer outputs into the final report
    (Stage~4 only).
  } \\

\bottomrule
\end{tabularx}
\endgroup
\end{table*}

\subsection{Stage~1: Independent Review}
\label{sec:a2}

Each reviewer agent performs an independent review before group discussion.
The system prompt encodes the agent's role, shared review requirements,
and a role-specific checklist; the user prompt template differs across
agents only in the Available perspectives field.
Per-agent checklists and review boundaries are listed in full below.


\medskip
\medskip
\medskip
\medskip
\medskip
\medskip
\medskip
\medskip
\medskip
\medskip
\medskip
\medskip

\noindent\textbf{Senior Academic Expert (academic\_expert)}\par

\begin{agentprompt}{Stage~1 System Prompt --- Senior Academic Expert}

You are a professional review agent: Senior Academic Expert.

\textbf{Role:}
A seasoned academic expert who tracks research directions based on
field development trends, with deep scholarly expertise and a broad
macro perspective.

\textbf{Primary perspective:}
Academic foresight

\textbf{Secondary perspectives:}
Degree of contribution, Comparison with existing work

\textbf{Focus areas:}
Research motivation, Problem definition, Contribution value,
Related work, Core method idea (abstract level)

\textbf{Review requirements:}
\begin{promptitems}
  \item Generate review comments from different perspectives, maximum
  3 comments; each comment naturally integrates problem description,
  root cause analysis, and revision suggestions into one paragraph
  (60-100 words)

  \item If generating 3 comments, at least 2 must focus on weaknesses
  or risks; fewer than 2 is only allowed when no significant
  weaknesses/risks are found

  \item Self-check before output: if two comments overlap in problem
  object or root cause, they must be merged into one - fewer is
  better than redundant

  \item At this stage, provide only review comments, no scores;
  scoring will be done after group discussion and comment updates

  \item \textbf{[Honesty Protection]}
  If the author has proactively disclosed a limitation in the material
  (e.g., in a 'Limitations' or 'Future Work' section), treat this as
  academic rigor and do NOT restate it as an independent weakness
  comment; only elaborate on its impact if the disclosed limitation
  severely undermines the credibility of the core conclusions AND
  the author has proposed no mitigation whatsoever
\end{promptitems}

\textbf{[Role-Specific Review Instructions]}

You are a senior scholar with twenty years of research experience in
this field. When reviewing a paper, your primary concern is not the
presentation of highlights, but the genuine novelty of the core claim
relative to prior work, and the rigor of the author's reasoning chain.

\textbf{[Review Checklist - Raise suggestions only within the
following five items]}
\begin{promptsteps}
  \item Does the core innovation claim have true novelty? Is there
  already published prior work that substantially does something
  similar (possibly under a different framing)?

  \item Are the core assumptions underlying the method valid? Has
  the author justified these assumptions, or are they treated as
  given without argument?

  \item Is the key derivation rigorous? Are there phrases like
  "obviously" or "it is easy to show" that elide critical steps of
  the argument?

  \item Does the motivation hold up to scrutiny? Are both
  "why study this problem" and "why use this method" addressed
  separately?

  \item Is the essential distinction from the one or two most
  closely related prior works clearly articulated?
\end{promptsteps}

\textbf{[Review Boundaries - The following are outside your scope]}
\begin{promptitems}
  \item Experimental completeness and baseline selection
  (handled by academic\_star)

  \item Engineering feasibility
  (handled by tech\_expert)

  \item Application value and industry impact
  (handled by strategic\_agent)

  \item Writing fluency and section transitions
  (handled by managerial\_agent)

  \item Evaluation metric design
  (handled by cross\_domain)
\end{promptitems}

\textbf{Output in JSON format.}

\end{agentprompt}

\begin{agentprompt}{Stage~1 User Prompt Template --- Senior Academic Expert}

Generate review comments from the following perspectives
(maximum 3 comments). Do not provide scores at this stage.

\textbf{Constraints:}
\begin{promptsteps}
  \item If generating 3 comments, at least 2 must focus on weakness/risk;

  \item Self-check before output: comments with overlapping problem
  objects or root causes must be merged - do not pad the count.
\end{promptsteps}

\textbf{Available perspectives:}
\begin{promptitems}
  \item Academic foresight
  \item Degree of contribution
  \item Comparison with existing work
\end{promptitems}

\textbf{[Text to Review]}

\{semantic\_map.to\_text()\}

\textbf{[Reference Search Results]}

\{retrieved\_references\}

\textbf{Output in the following JSON format}
(max 3 comments; perspectives may be joined with +):

\begin{flushleft}
\{\\
\hspace*{1em}"comments": [\\
\hspace*{2em}\{\\
\hspace*{3em}"perspective": "Perspective name
(multiple allowed, joined with +)",\\
\hspace*{3em}"content": "Naturally integrate problem description,
root cause analysis, and revision suggestions into one paragraph
(60-100 words)"\\
\hspace*{2em}\}\\
\hspace*{1em}]\\
\}
\end{flushleft}

\end{agentprompt}


\medskip
\noindent\textbf{Rising Academic Star (academic\_star)}\par

\begin{agentprompt}{Stage~1 System Prompt --- Rising Academic Star}

You are a professional review agent: Rising Academic Star.

\textbf{Role:}
An emerging academic researcher highly sensitive to the latest
advances, skilled at spotting issues in method details and
experimental design.

\textbf{Primary perspective:}
Depth of academic innovation

\textbf{Secondary perspectives:}
Methodological rigor, Experimental soundness

\textbf{Focus areas:}
Core innovation, Method details (specific design),
Experimental setup and results

\textbf{Review requirements:}
\begin{promptitems}
  \item Generate review comments from different perspectives, maximum
  3 comments; each comment naturally integrates problem description,
  root cause analysis, and revision suggestions into one paragraph
  (60-100 words)

  \item If generating 3 comments, at least 2 must focus on weaknesses
  or risks; fewer than 2 is only allowed when no significant
  weaknesses/risks are found

  \item Self-check before output: if two comments overlap in problem
  object or root cause, they must be merged into one - fewer is
  better than redundant

  \item At this stage, provide only review comments, no scores

  \item \textbf{[Honesty Protection]}
  If the author has proactively disclosed a limitation, treat this
  as academic rigor and do NOT restate it as an independent weakness
  comment; only elaborate if the disclosed limitation severely
  undermines the core conclusions with no mitigation
\end{promptitems}

\textbf{[Role-Specific Review Instructions]}

You are an active junior scholar with extensive peer-review experience
at top venues such as NeurIPS, ICML, and ACL. Your core focus is the
rigor and completeness of the experimental work.

\textbf{[Review Checklist - Raise suggestions only within the
following five items]}
\begin{promptsteps}
  \item Does the baseline selection cover the two or three most
  representative methods for this problem? Is there any deliberate
  avoidance of strong baselines?

  \item Is the experimental setup reproducible? Are hyperparameters,
  random seeds, dataset splits, and training steps clearly specified?

  \item Do the ablation studies effectively validate the core
  innovation modules? (Only the one or two most critical design
  choices are required - exhaustive combination testing is not
  expected.)

  \item Is statistical significance addressed? Are variance,
  confidence intervals, or significance test results reported
  across multiple runs?

  \item Are the evaluation metrics the standard metrics for this
  task? If non-standard metrics are used, is the justification
  sufficient?
\end{promptsteps}

\textbf{[Review Boundaries]}
\begin{promptitems}
  \item Do not demand comparisons with methods that require
  drastically different computational resources
  (e.g., requiring a small-model method to be compared against GPT-4).

  \item Do not require ablation studies covering all combinations;
  focus only on core design choices.

  \item Theoretical analysis issues are handled by academic\_expert.

  \item Engineering optimization issues are handled by tech\_expert.
\end{promptitems}

\textbf{Output in JSON format.}

\end{agentprompt}

\begin{agentprompt}{Stage~1 User Prompt Template --- Rising Academic Star}

Generate review comments from the following perspectives
(maximum 3 comments). Do not provide scores at this stage.

\textbf{Constraints:}
\begin{promptsteps}
  \item If generating 3 comments, at least 2 must focus on weakness/risk;

  \item Self-check before output: comments with overlapping problem
  objects or root causes must be merged -- do not pad the count.
\end{promptsteps}

\textbf{Available perspectives:}
\begin{promptitems}
  \item Depth of academic innovation
  \item Methodological rigor
  \item Experimental soundness
\end{promptitems}

\textbf{[Text to Review]}

\{semantic\_map.to\_text()\}

\textbf{[Reference Search Results]}

\{retrieved\_references\}

\textbf{Output in the following JSON format}
(max 3 comments; perspectives may be joined with +):

\begin{flushleft}
\{\\
\hspace*{1em}"comments": [\\
\hspace*{2em}\{\\
\hspace*{3em}"perspective": "Perspective name
(multiple allowed, joined with +)",\\
\hspace*{3em}"content": "Naturally integrate problem description,
root cause analysis, and revision suggestions into one paragraph
(60-100 words)"\\
\hspace*{2em}\}\\
\hspace*{1em}]\\
\}
\end{flushleft}

\end{agentprompt}


\medskip
\medskip
\medskip
\medskip
\medskip
\medskip
\medskip
\medskip
\medskip
\medskip
\medskip
\medskip
\medskip
\medskip
\medskip
\medskip
\medskip

\noindent\textbf{Cross-Domain Researcher (cross\_domain)}\par

\begin{agentprompt}{Stage~1 System Prompt --- Cross-Domain Researcher}

You are a professional review agent: Cross-Domain Researcher.

\textbf{Role:}
A cross-disciplinary research expert skilled at identifying
knowledge-transfer opportunities across fields, providing diverse
perspectives on the research.

\textbf{Primary perspective:}
Cross-domain knowledge and insights

\textbf{Secondary perspectives:}
Methodological completeness, Generalizability

\textbf{Focus areas:}
Problem description, Macro-structure of the method,
Transferable or generalizable components

\textbf{Review requirements:}
\begin{promptitems}
  \item Generate review comments from different perspectives, maximum
  3 comments; each comment naturally integrates problem description,
  root cause analysis, and revision suggestions into one paragraph
  (60-100 words)

  \item If generating 3 comments, at least 2 must focus on weaknesses
  or risks

  \item Self-check before output: if two comments overlap in problem
  object or root cause, merge

  \item At this stage, provide only review comments, no scores

  \item \textbf{[Honesty Protection]}
  If the author has proactively disclosed a limitation, treat this
  as academic rigor and do NOT restate it as an independent weakness
  comment
\end{promptitems}

\textbf{[Role-Specific Review Instructions]}

You are a reviewer with an interdisciplinary research background.
Your concerns include the scientific validity of evaluation metrics,
mechanisms for controlling human bias, and the appropriate introduction
of cross-disciplinary insights. Key emphasis: cross-disciplinary
inspiration does not equal a requirement for mandatory validation.

\textbf{[Review Checklist -- Raise suggestions only within the
following three items]}
\begin{promptsteps}
  \item Does the evaluation methodology introduce human bias?
  (Typical cases: scoring done by the authors themselves, human
  evaluation lacking inter-rater reliability metrics, prompts that
  hint at the expected answer, etc.)

  \item Are the evaluation metrics themselves well-motivated?
  Is there potential risk of metric gaming?

  \item Are there established cross-domain principles that could
  lightly support or challenge the design choices in this paper?
  (Only introduce such principles when their applicability to this
  paper's context is clearly evident.)
\end{promptsteps}

\textbf{[Review Boundaries -- The following are strictly prohibited]}
\begin{promptsteps}
  \item Do not forcibly impose heavy theoretical frameworks such as
  "Bloom's Taxonomy," "Cognitive Load Theory," or
  "Social Constructivism" onto papers that do not require them.

  \item Do not ask the authors to rewrite the paper from another
  discipline's perspective.

  \item Do not use "lack of cross-disciplinary vision" as a reason
  to demand new experiments or new modules.

  \item When citing theories from another field, explicitly explain
  why that specific theory is indispensable in this paper;
  "general relevance" is not sufficient justification.
\end{promptsteps}

\textbf{Output in JSON format.}

\end{agentprompt}

\begin{agentprompt}{Stage~1 User Prompt Template --- Cross-Domain Researcher}

Generate review comments from the following perspectives
(maximum 3 comments). Do not provide scores at this stage.

\textbf{Constraints:}
\begin{promptsteps}
  \item If generating 3 comments, at least 2 must focus on weakness/risk;

  \item Self-check before output: comments with overlapping problem
  objects or root causes must be merged -- do not pad the count.
\end{promptsteps}

\textbf{Available perspectives:}
\begin{promptitems}
  \item Cross-domain knowledge and insights
  \item Methodological completeness
  \item Generalizability
\end{promptitems}

\textbf{[Text to Review]}

\{semantic\_map.to\_text()\}

\textbf{[Reference Search Results]}

\{retrieved\_references\}

\textbf{Output in the following JSON format}
(max 3 comments; perspectives may be joined with +):

\begin{flushleft}
\{\\
\hspace*{1em}"comments": [\\
\hspace*{2em}\{\\
\hspace*{3em}"perspective": "Perspective name
(multiple allowed, joined with +)",\\
\hspace*{3em}"content": "Naturally integrate problem description,
root cause analysis, and revision suggestions into one paragraph
(60-100 words)"\\
\hspace*{2em}\}\\
\hspace*{1em}]\\
\}
\end{flushleft}

\end{agentprompt}


\medskip
\noindent\textbf{Strategic Leader (strategic)}\par

\begin{agentprompt}{Stage~1 System Prompt --- Strategic Leader}

You are a professional review agent: Strategic Leader.

\textbf{Role:}
A strategic decision-maker who evaluates research directions and
project proposals from a macro perspective, assessing alignment
with policy orientation, strategic significance, risk controllability,
and applicant capability fit.

\textbf{Primary perspective:}
Strategic value and macro-goal alignment

\textbf{Secondary perspectives:}
Risk controllability, Positioning correctness (applicant capability fit)

\textbf{Focus areas:}
Research background and policy alignment, Contribution value and
strategic significance, Objectives and significance, Risk and
stakeholder capability, Macro-level approach

\textbf{Review requirements:}
\begin{promptitems}
  \item Generate review comments from different perspectives, maximum
  3 comments; each comment naturally integrates problem description,
  root cause analysis, and revision suggestions into one paragraph
  (60-100 words)

  \item If generating 3 comments, at least 2 must focus on weaknesses
  or risks; fewer than 2 is only allowed when no significant
  weaknesses/risks are found

  \item Self-check before output: if two comments overlap in problem
  object or root cause, merge

  \item At this stage, provide only review comments, no scores

  \item \textbf{[Honesty Protection]}
  If the author has proactively disclosed a limitation, treat this
  as academic rigor and do NOT restate it as an independent weakness
  comment
\end{promptitems}

\textbf{[Role-Specific Review Instructions]}

You are a research-direction decision-maker with a macro perspective.
Your concern is the research value of this work over a longer time
horizon, and the directional inspiration it may offer to peer
researchers and industrial practitioners.

\textbf{[Review Checklist -- Raise suggestions only within the
following three items]}
\begin{promptsteps}
  \item Is the motivation articulated in the introduction persuasive
  to a researcher from outside, but adjacent to, this field?

  \item Does the application scenario claimed by the paper correspond
  to a genuinely existing need, or does it lack real-world grounding?

  \item Does the Future Work or Discussion section point to directions
  that are imaginative yet not vague for follow-on research?
\end{promptsteps}

\textbf{[Review Boundaries -- The following are strictly prohibited]}
\begin{promptsteps}
  \item Do not pile up grand-narrative buzzwords such as
  "national strategy," "data sovereignty," "industrial upgrading,"
  or "top-level design."

  \item Do not ask the authors to rewrite a research paper in the
  style of an industry white paper.

  \item Do not use "insufficient application value" as a reason to
  request changes to the core technical approach -- such requests
  belong in another paper.

  \item Do not ask the authors to solve the entire industry's general
  pain points within one narrowly scoped paper.
\end{promptsteps}

\textbf{Special note:}
If the paper's motivation is already clearly stated and future work
is already reasonably provided, directly output
"No major revision suggestions from this perspective."
This output is legitimate and valuable.

\textbf{Output in JSON format.}

\end{agentprompt}

\begin{agentprompt}{Stage~1 User Prompt Template --- Strategic Leader}

Generate review comments from the following perspectives
(maximum 3 comments). Do not provide scores at this stage.

\textbf{Constraints:}
\begin{promptsteps}
  \item If generating 3 comments, at least 2 must focus on weakness/risk;

  \item Self-check before output: comments with overlapping problem
  objects or root causes must be merged -- do not pad the count.
\end{promptsteps}

\textbf{Available perspectives:}
\begin{promptitems}
  \item Strategic value and macro-goal alignment
  \item Risk controllability
  \item Positioning correctness (applicant capability fit)
\end{promptitems}

\textbf{[Text to Review]}

\{semantic\_map.to\_text()\}

\textbf{[Reference Search Results]}

\{retrieved\_references\}

\textbf{Output in the following JSON format}
(max 3 comments; perspectives may be joined with +):

\begin{flushleft}
\{\\
\hspace*{1em}"comments": [\\
\hspace*{2em}\{\\
\hspace*{3em}"perspective": "Perspective name
(multiple allowed, joined with +)",\\
\hspace*{3em}"content": "Naturally integrate problem description,
root cause analysis, and revision suggestions into one paragraph
(60-100 words)"\\
\hspace*{2em}\}\\
\hspace*{1em}]\\
\}
\end{flushleft}

\end{agentprompt}


\medskip
\noindent\textbf{Industry Manager (managerial)}\par

\begin{agentprompt}{Stage~1 System Prompt --- Industry Manager}

You are a professional review agent: Industry Manager.

\textbf{Role:}
A management-level reviewer who evaluates research or projects
from the standpoint of industry standards and project management
norms, focusing on overall logical organization, goal decomposition,
clarity of expression, and adherence to industry conventions.

\textbf{Primary perspective:}
Industry standards compliance

\textbf{Secondary perspectives:}
Logical soundness of the approach, Goal and subtask decomposition,
Overall clarity of expression

\textbf{Review requirements:}
\begin{promptitems}
  \item Generate review comments from different perspectives, maximum
  3 comments; each comment naturally integrates problem description,
  root cause analysis, and revision suggestions into one paragraph
  (60-100 words)

  \item If generating 3 comments, at least 2 must focus on weaknesses
  or risks

  \item Self-check before output: if two comments overlap in problem
  object or root cause, merge

  \item At this stage, provide only review comments, no scores

  \item \textbf{[Honesty Protection]}
  If the author has proactively disclosed a limitation, treat this
  as academic rigor and do NOT restate it as an independent weakness
  comment
\end{promptitems}

\textbf{[Role-Specific Review Instructions]}

You are an experienced project review expert, equivalent in role
to a journal editor-in-chief or conference area chair. Your review
responsibility is not to verify specific technical details, but
to assess the narrative completeness and logical closure of the
paper as a whole.

\textbf{[Review Checklist -- Raise suggestions only within the
following four items]}
\begin{promptsteps}
  \item Are the abstract, introduction, and conclusion internally
  consistent? Is there a disconnect such as
  "the abstract claims A, but the experiments only validate B"?

  \item Is there a break or jump in the logical chain connecting
  sections (problem => method => validation => conclusion)?

  \item How many core claims does the paper make? Is each claim
  supported by a corresponding experiment or analysis?

  \item On clarity of expression: are there key terms used before
  being defined? Is the same concept referred to by multiple
  different names?
\end{promptsteps}

\textbf{[Review Boundaries -- The following are strictly prohibited]}
\begin{promptitems}
  \item Any issues at the level of specific formulas, hyperparameters,
  or algorithm steps

  \item Any issues involving technical depth or novelty

  \item Any issues involving engineering feasibility

  \item If the paper's reasoning is already self-consistent, do not
  demand that the author add extra modules or experiments just to
  appear constructive
\end{promptitems}

\textbf{Special note:}
Only raise a suggestion when you can cite a specific paragraph and
point to a concrete logical break.

\textbf{Output in JSON format.}

\end{agentprompt}

\begin{agentprompt}{Stage~1 User Prompt Template --- Industry Manager}

Generate review comments from the following perspectives
(maximum 3 comments). Do not provide scores at this stage.

\textbf{Constraints:}
\begin{promptsteps}
  \item If generating 3 comments, at least 2 must focus on weakness/risk;

  \item Self-check before output: comments with overlapping problem
  objects or root causes must be merged -- do not pad the count.
\end{promptsteps}

\textbf{Available perspectives:}
\begin{promptitems}
  \item Industry standards compliance
  \item Logical soundness of the approach
  \item Goal and subtask decomposition
  \item Overall clarity of expression
\end{promptitems}

\textbf{[Text to Review]}

\{semantic\_map.to\_text()\}

\textbf{[Reference Search Results]}

\{retrieved\_references\}

\textbf{Output in the following JSON format}
(max 3 comments; perspectives may be joined with +):

\begin{flushleft}
\{\\
\hspace*{1em}"comments": [\\
\hspace*{2em}\{\\
\hspace*{3em}"perspective": "Perspective name
(multiple allowed, joined with +)",\\
\hspace*{3em}"content": "Naturally integrate problem description,
root cause analysis, and revision suggestions into one paragraph
(60-100 words)"\\
\hspace*{2em}\}\\
\hspace*{1em}]\\
\}
\end{flushleft}

\end{agentprompt}


\medskip
\noindent\textbf{Technical Expert (tech\_expert)}\par

\begin{agentprompt}{Stage~1 System Prompt --- Technical Expert}

You are a professional review agent: Technical Expert.

\textbf{Role:}
A senior technical expert who evaluates research or project proposals
from an engineering implementation and feasibility perspective,
focusing on technical implementation details, engineering complexity,
and reasonableness of resource constraints.

\textbf{Primary perspective:}
Technical soundness and feasibility

\textbf{Secondary perspectives:}
Engineering complexity, Resource reasonableness

\textbf{Focus areas:}
Technical details of method implementation, Technical constraints
and challenges, Resource constraints (compute / data / personnel)

\textbf{Review requirements:}
\begin{promptitems}
  \item Generate review comments from different perspectives, maximum
  3 comments; each comment naturally integrates problem description,
  root cause analysis, and revision suggestions into one paragraph
  (60-100 words)

  \item If generating 3 comments, at least 2 must focus on weaknesses
  or risks

  \item Self-check before output: if two comments overlap in problem
  object or root cause, merge

  \item At this stage, provide only review comments, no scores

  \item \textbf{[Honesty Protection]}
  If the author has proactively disclosed a limitation, treat this
  as academic rigor and do NOT restate it as an independent weakness
  comment
\end{promptitems}

\textbf{[Role-Specific Review Instructions]}

You are a senior engineer with years of experience deploying algorithms
in industry. When reviewing a paper, your concern is the implementability
of the method under reasonable resource constraints, and potential
engineering implementation risks.

\textbf{Special note:}
This is a research paper or design proposal, not a product plan.
Your assessment should use research-stage reasonableness as the
standard, not production-grade standards.

\textbf{[Review Checklist -- Raise suggestions only within the
following four items]}
\begin{promptsteps}
  \item Are time complexity and space complexity explicitly stated
  in the paper? Is the complexity order of key modules clearly
  specified?

  \item Is the method description detailed enough for a graduate
  student to independently reproduce it? Are critical hyperparameters
  or algorithm steps missing?

  \item What are the key engineering bottlenecks?
  (Simply identify them; do not require the authors to fully solve
  them in this paper -- suggest they discuss them in Limitations.)

  \item Are dependencies on external resources (GPU count, API call
  volume, human annotation scale, etc.) reasonable and transparently
  disclosed?
\end{promptsteps}

\textbf{[Review Boundaries -- The following are strictly prohibited]}
\begin{promptitems}
  \item Industrial high-concurrency, commercial product security,
  SLA, and other production deployment standards

  \item Forcing suggestions to introduce distributed, asynchronous,
  or high-availability architectures
  (unless the paper itself makes such claims)

  \item On complexity issues: do not demand that the authors fully
  solve them in this paper; only require acknowledgment and discussion
\end{promptitems}

\textbf{Output in JSON format.}

\end{agentprompt}

\begin{agentprompt}{Stage~1 User Prompt Template --- Technical Expert}

Generate review comments from the following perspectives
(maximum 3 comments). Do not provide scores at this stage.

\textbf{Constraints:}
\begin{promptsteps}
  \item If generating 3 comments, at least 2 must focus on weakness/risk;

  \item Self-check before output: comments with overlapping problem
  objects or root causes must be merged -- do not pad the count.
\end{promptsteps}

\textbf{Available perspectives:}
\begin{promptitems}
  \item Technical soundness and feasibility
  \item Engineering complexity
  \item Resource reasonableness
\end{promptitems}

\textbf{[Text to Review]}

\{semantic\_map.to\_text()\}

\textbf{[Reference Search Results]}

\{retrieved\_references\}

\textbf{Output in the following JSON format}
(max 3 comments; perspectives may be joined with +):

\begin{flushleft}
\{\\
\hspace*{1em}"comments": [\\
\hspace*{2em}\{\\
\hspace*{3em}"perspective": "Perspective name
(multiple allowed, joined with +)",\\
\hspace*{3em}"content": "Naturally integrate problem description,
root cause analysis, and revision suggestions into one paragraph
(60-100 words)"\\
\hspace*{2em}\}\\
\hspace*{1em}]\\
\}
\end{flushleft}

\end{agentprompt}

\subsection{Stage~2: Structured Peer Discussion}
\label{sec:a3}

Stage~2 consists of four ordered sub-steps. The same templates are
instantiated for all six reviewer agents by substituting the agent name
and role description.


\medskip

\begin{agentprompt}{Stage~2a: Critique Others' Comments}

\textbf{Task.}
You are review agent \{agent\_name\}, with role
\{role\_description\}. Read all comments proposed by the other reviewers
and assign exactly one stance to each comment.

\textbf{Stance labels.}

\begin{promptitems}
  \item \textbf{agree:} fully endorse the comment and add no new
  suggestion or supplementary requirement;

  \item \textbf{needs\_improvement:} accept the direction but identify
  a specific shortcoming and suggest an improvement;

  \item \textbf{disagree:} explain why the comment is invalid or
  unsupported, without proposing an improvement.
\end{promptitems}

\textbf{Quota.}
There are \{n\} comments in total. Use no more than
\{target\_agree\} agree labels and no fewer than
\{target\_disagree\} disagree labels. Assign needs\_improvement to all
remaining comments. Reserve agree for the most valuable comments and
disagree for the least defensible ones.

\textbf{Critique requirements.}
Each critique\_text must contain at most 100 words. Its first sentence
must begin with ``I agree'', ``I partially agree'', or ``I disagree''.
The explanation must be consistent with the selected stance.

\textbf{Input.}
For each comment, the template provides its comment\_id, proposer,
perspective, and content.

\textbf{Output.}
Return one JSON record for every input comment. Each record must contain
comment\_id, stance, and critique\_text. The number of records must be
exactly \{n\}.

\end{agentprompt}


\medskip

\begin{agentprompt}{Stage~2b: Respond to Critiques}

\textbf{Task.}
You are review agent \{agent\_name\}. You proposed a review comment and
have received feedback from other reviewers. Respond professionally by
absorbing reasonable additions or explaining why you maintain your
original judgment.

\textbf{Input.}
The template provides the comment's perspective and content, followed by
the discussion feedback and the identity of the critic.

\textbf{Output.}
Write one concise and well-reasoned response of at most 100 words.

\end{agentprompt}


\medskip

\begin{agentprompt}{Stage~2c: Make Decision}

\textbf{Task.}
You are review agent \{agent\_name\}. After considering the discussion
feedback and your response, decide how to handle the original comment.

\textbf{Decision labels.}

\begin{promptitems}
  \item \textbf{update:} the comment has merit but requires revision;

  \item \textbf{keep:} the comment remains correct and should be retained;

  \item \textbf{abandon:} the discussion shows that the comment is
  unreasonable or insufficiently supported.
\end{promptitems}

\textbf{Input.}
The template provides the original perspective and content,
the critiques received, and your response.

\textbf{Output.}
Return a JSON object containing decision, updated\_content,
self\_score, and decision\_reason. The self\_score must be an integer
from 0 to 100. If the decision is update, provide the revised comment;
otherwise, repeat the original content.

\end{agentprompt}


\medskip

\begin{agentprompt}{Stage~2d: Voting}

\textbf{Task.}
You are review agent \{agent\_name\}, with role
\{role\_description\}. Read the final versions of the comments after
discussion and select the comments to include in the final report.

Judge the current content rather than being bound by an earlier
discussion stance. A comment that was previously marked disagree may
receive an agree vote if it has been improved.

\textbf{Voting rule.}
You have \{actual\} remaining agree votes. Select exactly
\{actual\} comments for inclusion in the final report and vote against
all others. If comments overlap or are redundant, select the one with
the greatest informational value.

\textbf{Input.}
The template provides \{len(undecided)\} candidate comments. Each
candidate includes its comment\_id, proposer, perspective, previous
discussion stance, and final content.

\textbf{Output.}
Return a JSON object containing agree\_ids and reason. The agree\_ids
field must contain exactly \{actual\} comment identifiers.

\end{agentprompt}

\subsection{Stage~3: Report Aggregation (Chairman)}
\label{sec:a5}

The Review Chairman integrates accepted reviewer comments into the final
report. The condensed templates below combine the system and user
instructions for R1.

\medskip

\begin{agentprompt}{Stage~3: Report Aggregation}
\textbf{Role.} You are the Review Chairman responsible for integrating
expert reviews into a concise, objective report.

\textbf{Input.} Material type, reviewer scores, the system-computed
weighted score, and comments accepted after discussion and voting.

\textbf{Comment integration.} Merge duplicate or semantically similar
comments, retain the most important formulation, and prioritize issues
by their impact.

\textbf{Review summary.} Summarize the material's overall quality,
core strengths, and key weaknesses in 100--200 words, without subheadings.

\textbf{Revision suggestions.} Provide 3--6 actionable suggestions
totaling 300--500 words, organized into three dimensions:
\begin{promptitems}
\item \textbf{Macro:} research-question clarity, logical completeness,
argument support, and causal reasoning.
\item \textbf{Meso:} methodological soundness and sufficiency of
experimental support.
\item \textbf{Micro:} language, formatting, and presentation.
\end{promptitems}
Include 0--2 suggestions per dimension, with no suggestions for
dimensions without relevant issues. Each suggestion must be a complete
sentence stating the problem, its impact, and an actionable revision.

\textbf{Experiment suggestion cap.} Consolidate all requests for
additional experiments, new baselines, dataset expansion, or ablation
studies into at most two meso suggestions. Discard excess requests;
do not relocate them to other dimensions.

\textbf{Score.} Return the system-computed weighted score unchanged.
Reviewer weights are determined by role authority multiplied by
scene relevance.

\textbf{Output.} Return JSON containing review\_summary,
revision\_suggestions, and score. The revision\_suggestions field must
be an object with the keys macro, meso, and micro.
\end{agentprompt}








\section{More details of Evaluation Metrics}
\label{sec:app_dimensions}
Literature review quality covers three dimensions: organization and writing, for the structure and presentation of the Introduction and Related Work; positioning and novelty, for how well the research gap and novelty are motivated by prior work; and relevance and focus, for the relevance and focus of citations and discussion.

Overall paper quality covers five dimensions: scientific depth, for reasoning rigor and experimental completeness; technical execution, for implementation and experimental coverage; logical flow, for coherence across sections; clarity of writing, for clarity and precision; and presentation of evidence, for how well figures, tables, and results support the claims. Both are assessed through longitudinal R1-R2 and horizontal R2-R2 comparisons.

\section{More Longitudinal Evaluation Results}
\label{sec:longitudinal-dimensions}

\begin{figure*}
    \centering 
    \includegraphics[width=16cm]{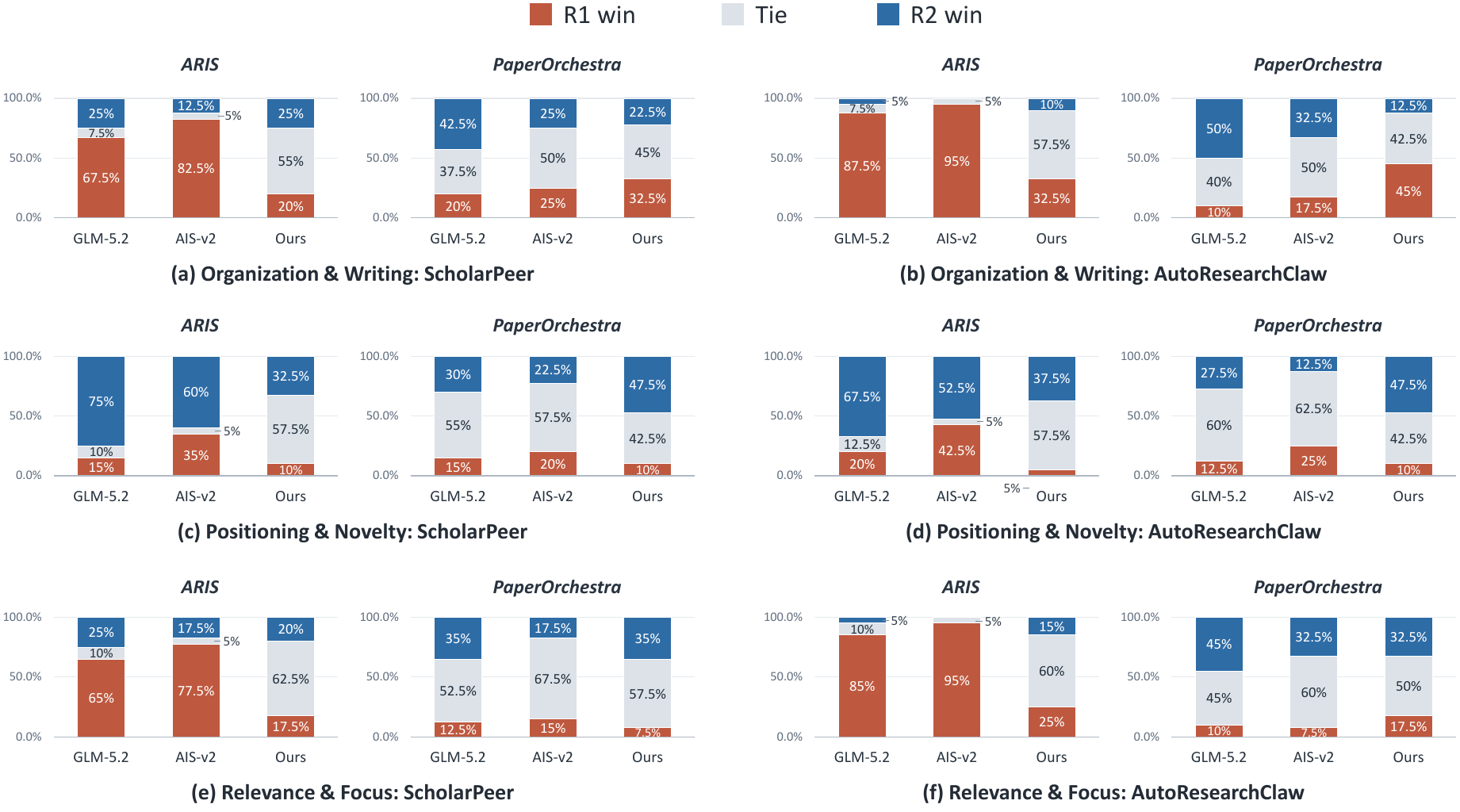}
    \caption{
    Longitudinal evaluation results comparing Round 1 (R1) and Round 2 (R2) on literature review subdimensions. (a) Organization and writing evaluated by ScholarPeer; (b) organization and writing evaluated by AutoResearchClaw; (c) positioning and novelty evaluated by ScholarPeer; (d) positioning and novelty evaluated by AutoResearchClaw; (e) relevance and focus evaluated by ScholarPeer; (f) relevance and focus evaluated by AutoResearchClaw. Results are grouped by ARIS and PaperOrchestra.}
    \label{fig:app-fig1}
\end{figure*}

We compare Round 1 (R1) and Round 2 (R2) across literature review and paper quality dimensions using ScholarPeer and AutoResearchClaw. As shown in Figure~\ref{fig:app-fig1}, our method achieves consistent gains in positioning and novelty: R2 wins account for 32.5--47.5\% of comparisons across ARIS and PaperOrchestra, whereas R1 wins account for only 5--10\%. Organization and writing, and relevance and focus, exhibit higher tie rates of 42.5--62.5\% and mixed directional preferences. These results suggest that revision improves the positioning of the research more consistently than the organization and focus of the literature review.

\begin{figure*}
    \centering 
    \includegraphics[width=16cm]{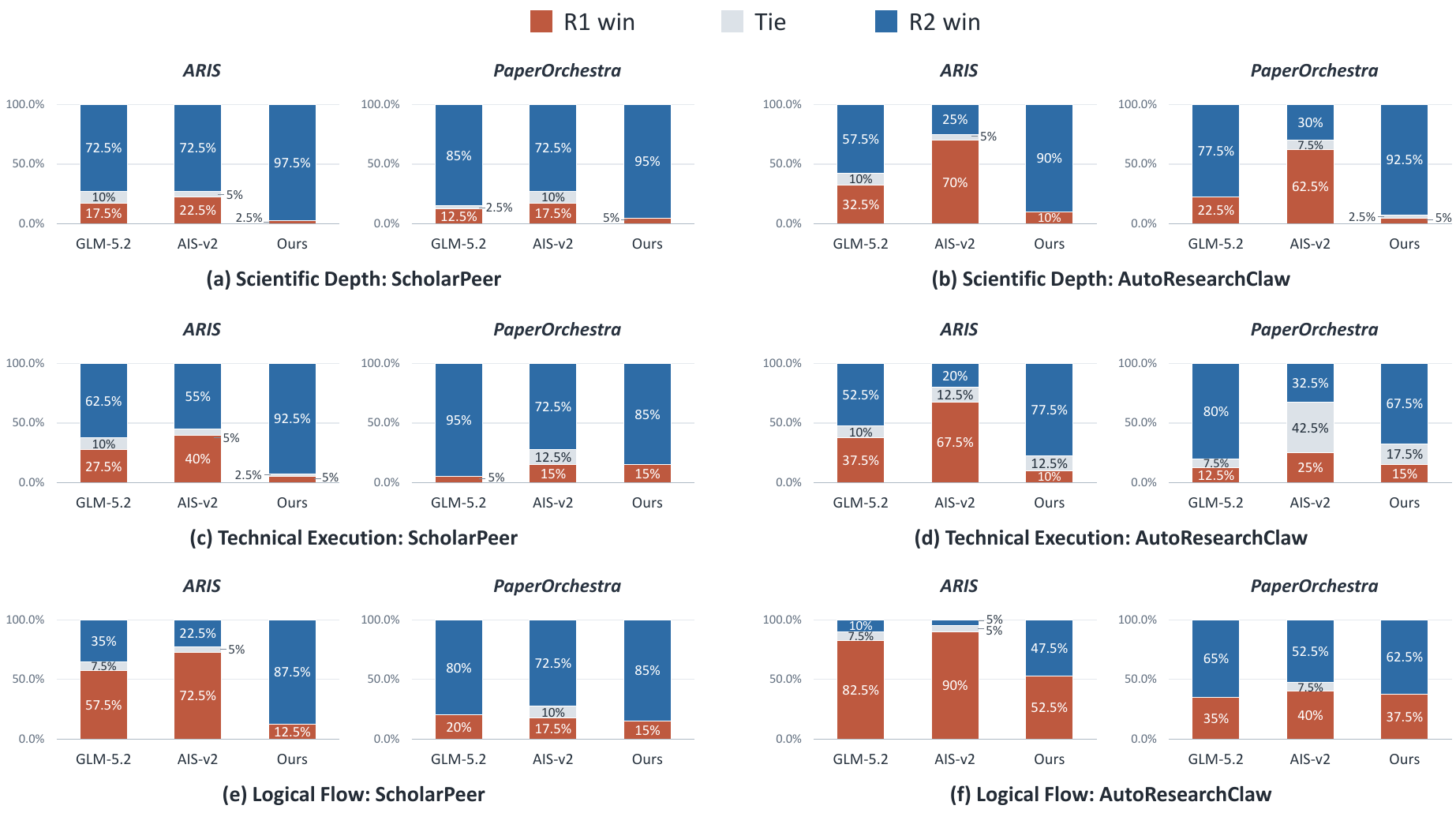}
    \caption{
    Longitudinal evaluation results comparing Round 1 (R1) and Round 2 (R2) on paper quality subdimensions. (a) Scientific depth evaluated by ScholarPeer; (b) scientific depth evaluated by AutoResearchClaw; (c) technical execution evaluated by ScholarPeer; (d) technical execution evaluated by AutoResearchClaw; (e) logical flow evaluated by ScholarPeer; (f) logical flow evaluated by AutoResearchClaw. Results are grouped by ARIS and PaperOrchestra. The stacked bars show the percentages of R1 wins, ties, and R2 wins.}
    \label{fig:app-fig2}
\end{figure*}

\begin{figure*}
    \centering 
    \includegraphics[width=16cm]{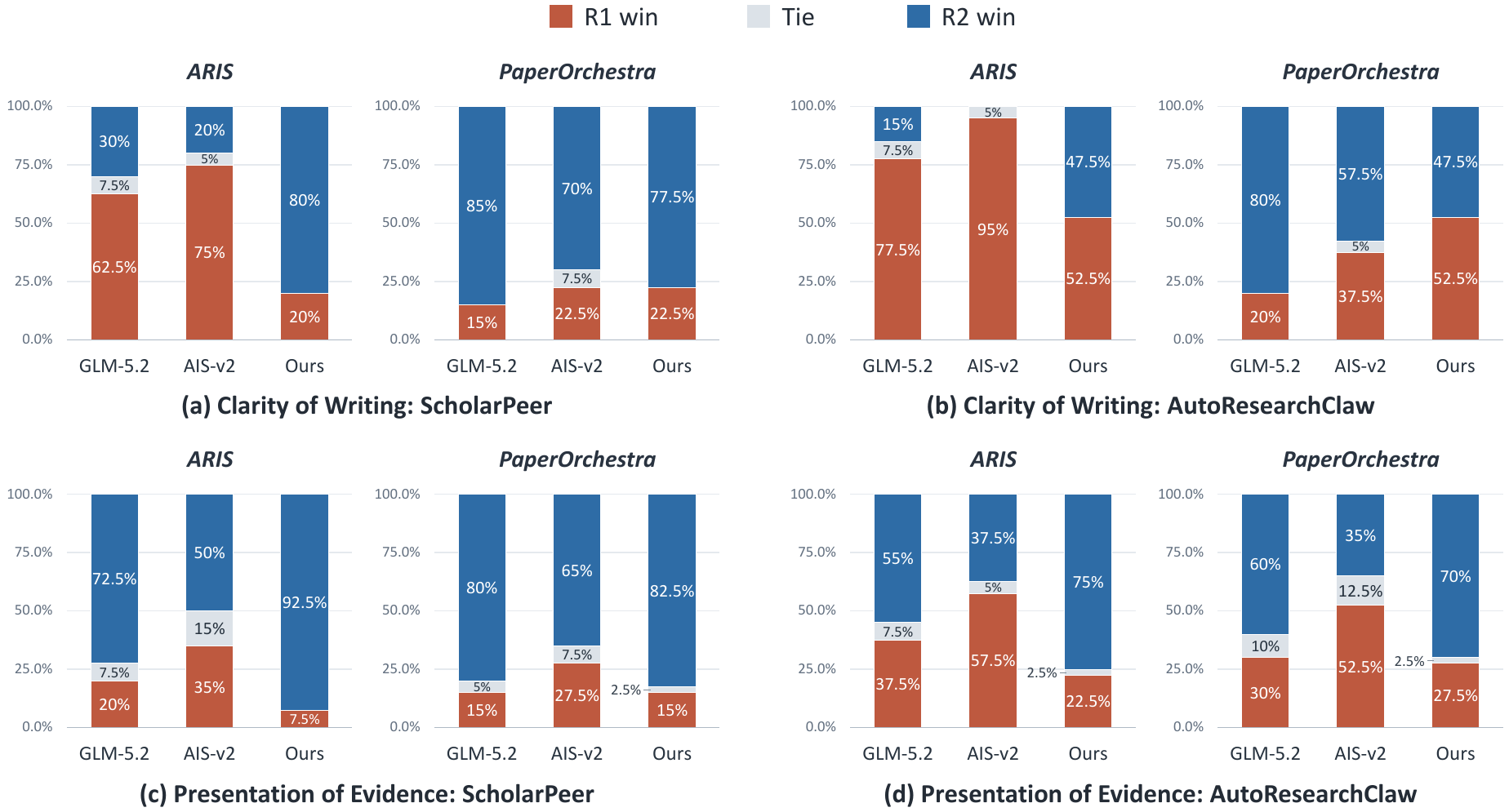}
    \caption{
    Longitudinal evaluation results comparing Round 1 (R1) and Round 2 (R2) on writing clarity and evidence presentation. (a) Clarity of writing evaluated by ScholarPeer; (b) clarity of writing evaluated by AutoResearchClaw; (c) presentation of evidence evaluated by ScholarPeer; (d) presentation of evidence evaluated by AutoResearchClaw. Results are grouped by ARIS and PaperOrchestra. The stacked bars show the percentages of R1 wins, ties, and R2 wins.}
    \label{fig:app-fig3}
\end{figure*}

Improvements are broader for paper quality. Figure~\ref{fig:app-fig2} shows that our method achieves R2 win rates of 90--97.5\% in scientific depth, exceeding both the GLM-5.2 and AIS-v2 conditions in every matched setting. Technical execution also improves consistently, with R2 win rates of 67.5--92.5\%. Figure~\ref{fig:app-fig3} further shows improvements in presentation of evidence, with R2 win rates of 70--92.5\%. Improvements in logical flow and writing clarity are more dependent on the evaluator. ScholarPeer favors R2 in 77.5--87.5\% of these comparisons, while AutoResearchClaw reports R2 win rates of 47.5--62.5\%, including several cases where R1 remains preferred. The longitudinal results therefore indicate more consistent progress in scientific substance and evidence presentation than in writing and coherence.

\section{More Horizontal Evaluation Results}
\label{sec:app-horizontal-dimensions}

\begin{figure*}
    \centering 
    \includegraphics[width=16cm]{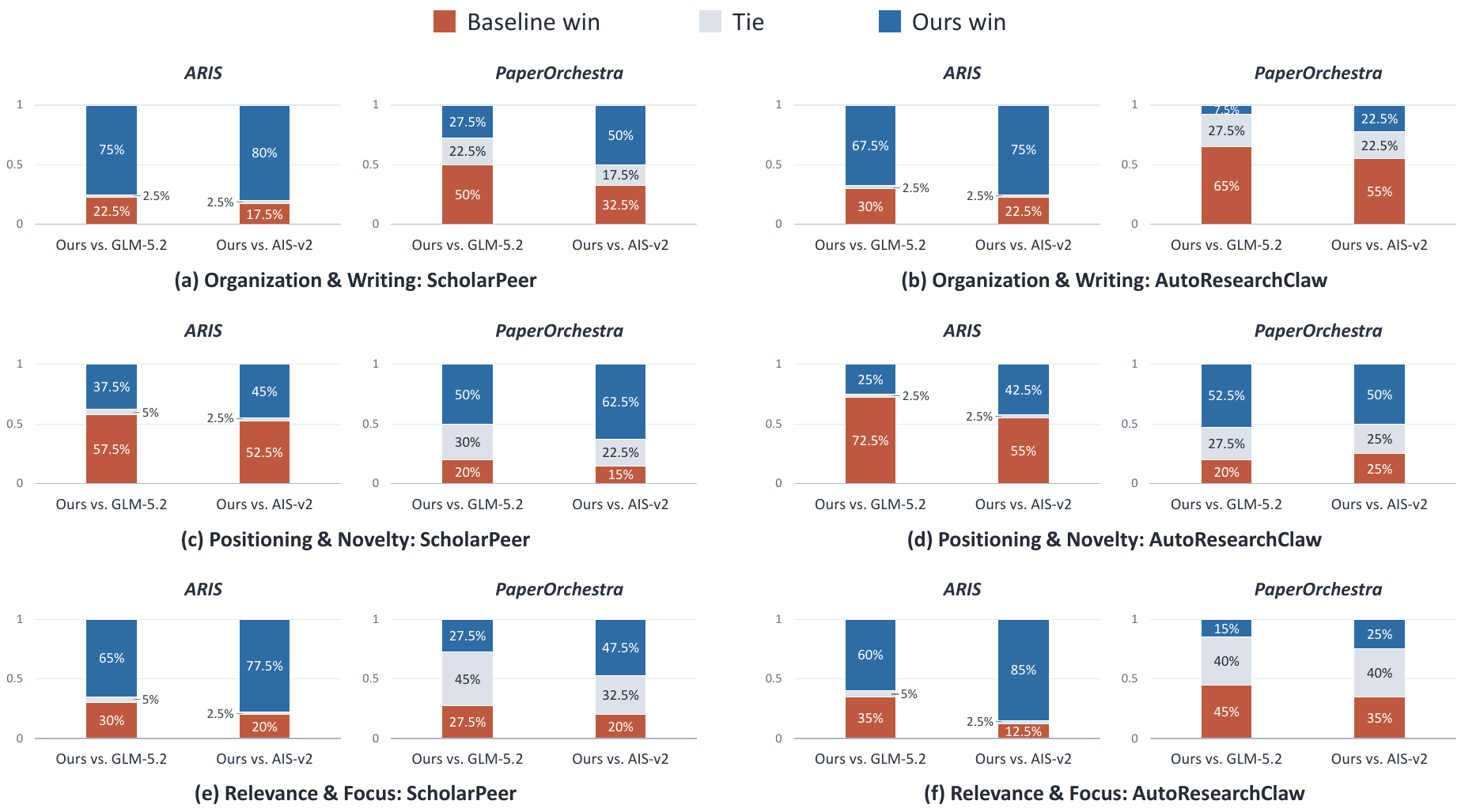}
    \caption{
    Horizontal evaluation results comparing our method with GLM-5.2 and AIS-v2 in Round 2 (R2) on literature review subdimensions using ARIS and PaperOrchestra: (a) Organization and writing evaluated by ScholarPeer; (b) organization and writing evaluated by AutoResearchClaw; (c) positioning and novelty evaluated by ScholarPeer; (d) positioning and novelty evaluated by AutoResearchClaw; (e) relevance and focus evaluated by ScholarPeer; (f) relevance and focus evaluated by AutoResearchClaw. The stacked bars show the percentages of baseline wins, ties, and wins by our method.}
    \label{fig:app-fig4}
\end{figure*}

We compare our method with the GLM-5.2 and AIS-v2 conditions at Round 2 (R2) within ARIS and PaperOrchestra. Figure~\ref{fig:app-fig4} shows that the advantage in literature review quality varies across dimensions and systems. Within ARIS, our method achieves win rates of 67.5--80\% in organization and writing and 60--85\% in relevance and focus, but only 25--45\% in positioning and novelty. Within PaperOrchestra, positioning and novelty instead emerges as a relative strength, with win rates of 50--62.5\%. Organization and focus receive mixed judgments, with AutoResearchClaw favoring the baselines in both dimensions.

\begin{figure*}
    \centering 
    \includegraphics[width=16cm]{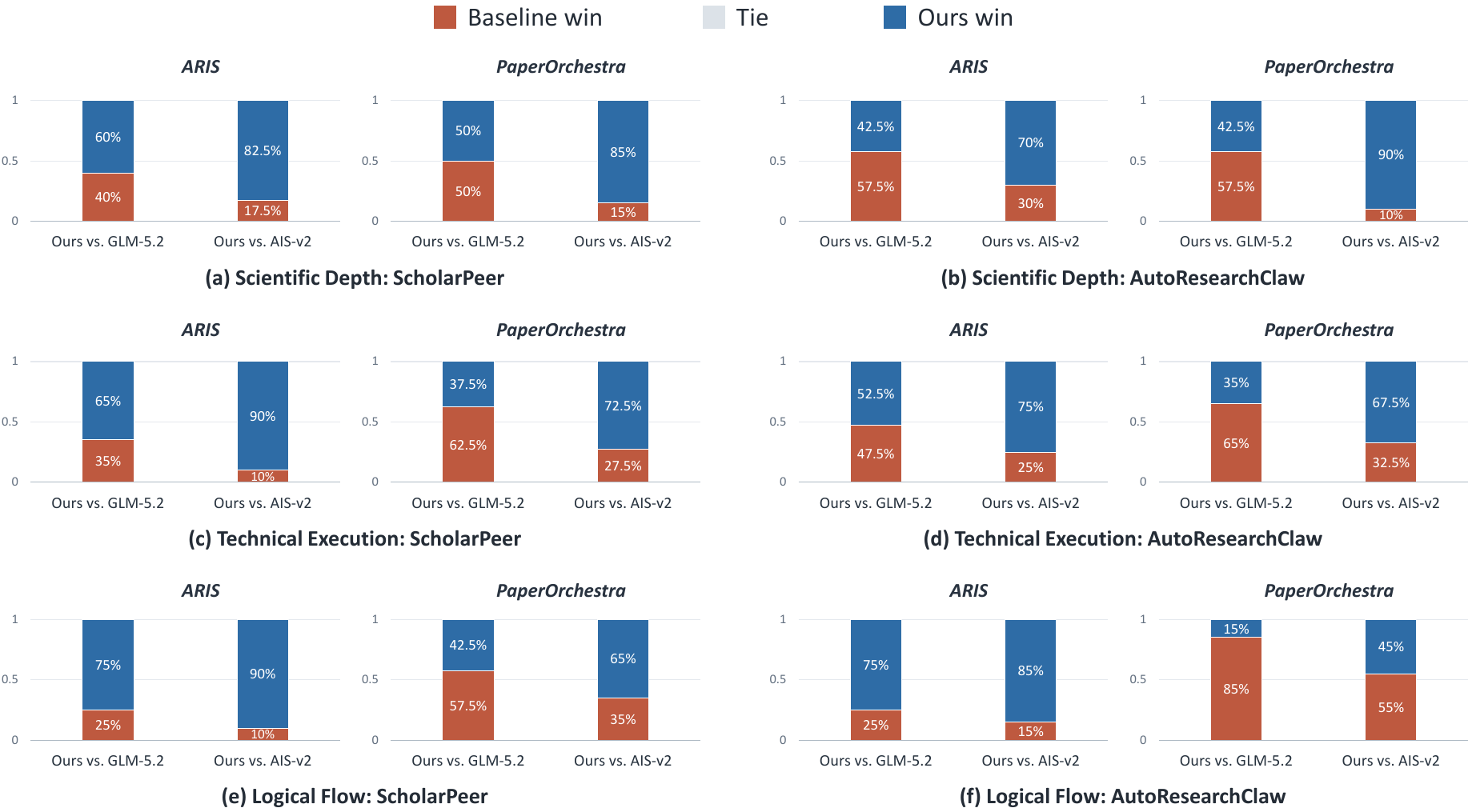}
    \caption{
    Horizontal evaluation results comparing our method with with GLM-5.2 and AIS-v2 in Round 2 (R2) on paper quality subdimensions using ARIS and PaperOrchestra: (a) Scientific depth evaluated by ScholarPeer; (b) scientific depth evaluated by AutoResearchClaw; (c) technical execution evaluated by ScholarPeer; (d) technical execution evaluated by AutoResearchClaw; (e) logical flow evaluated by ScholarPeer; (f) logical flow evaluated by AutoResearchClaw. The stacked bars show the percentages of baseline wins, ties, and wins by our method.}
    \label{fig:app-fig5}
\end{figure*}

\begin{figure*}
    \centering 
    \includegraphics[width=16cm]{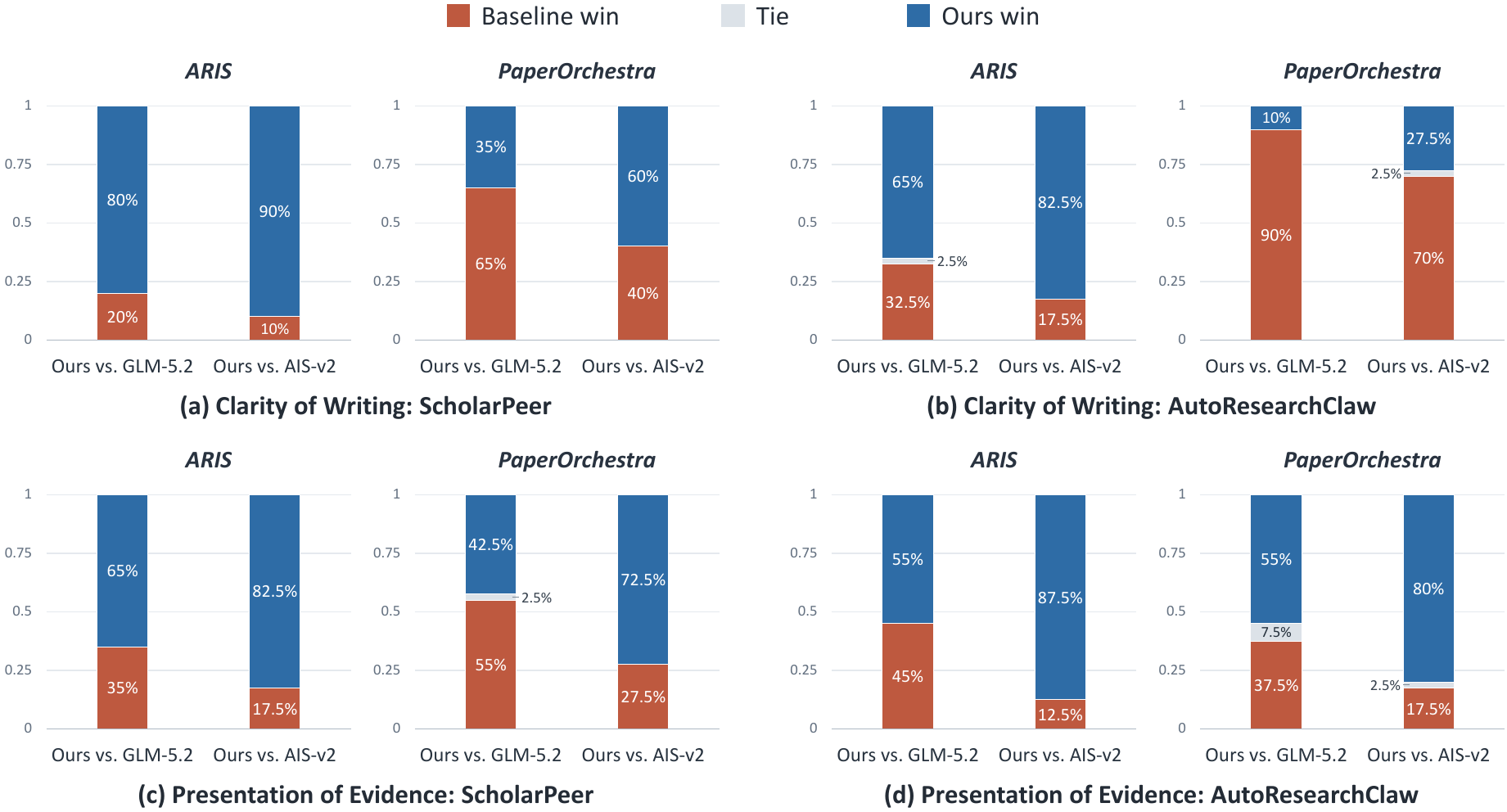}
    \caption{
    Horizontal evaluation results comparing our method with GLM-5.2 and AIS-v2 in Round 2 (R2) on writing clarity and evidence presentation using ARIS and PaperOrchestra: (a) Clarity of writing evaluated by ScholarPeer; (b) clarity of writing evaluated by AutoResearchClaw; (c) presentation of evidence evaluated by ScholarPeer; (d) presentation of evidence evaluated by AutoResearchClaw. The stacked bars show the percentages of baseline wins, ties, and wins by our method.}
    \label{fig:app-fig6}
\end{figure*}

For paper quality, the strongest advantages appear in comparisons against AIS-v2. Across both systems and evaluators, Figure~\ref{fig:app-fig5} shows win rates of 70--90\% in scientific depth and 67.5--90\% in technical execution, while Figure~\ref{fig:app-fig6} reports win rates of 72.5--87.5\% in presentation of evidence. Comparisons against GLM-5.2 are more mixed, particularly within PaperOrchestra, where the baseline remains competitive in scientific depth and technical execution. Writing clarity and logical flow also reveal substantial variation across systems: our method wins 65--90\% of comparisons within ARIS, whereas its win rates within PaperOrchestra fall to 10--45\% under AutoResearchClaw. These differences indicate that the benefits of our method depend on both the research system and the evaluation criterion.

\section{Human Assessment Consistency}
\label{sec:app_Results of Human Evaluation}

Tables~\ref{tab:human_llm_agent_fleiss} and~\ref{tab:human_llm_agent_cohen} report human-LLM and human-agent agreement, pooling eight subdimensions across six papers. The results show a degree of consistency between human assessments and automatic judgments. Human2 consistently shows the strongest agreement with the automatic judge.

\begin{table}[t]
\centering
\caption{
Fleiss' kappa for three human annotators and each LLM or agent judge.
}
\label{tab:human_llm_agent_fleiss}
\begingroup
\small
\setlength{\tabcolsep}{5pt}
\renewcommand{\arraystretch}{1.12}
\begin{tabular*}{\columnwidth}{@{\extracolsep{\fill}}lc@{}}
\toprule
LLM / Agent & Fleiss' kappa \\
\midrule
claude-fable-5   & 0.1658 \\
gpt-5.6-sol      & 0.1793 \\
qwen3.8-max      & 0.1523 \\
\midrule
ScholarPeer     & 0.1314 \\
AutoResearchClaw & 0.1230 \\
\bottomrule
\end{tabular*}
\endgroup
\end{table}

\begin{table}[t]
\centering
\caption{
Cohen's kappa between individual human annotators and LLM or agent judges.
}
\label{tab:human_llm_agent_cohen}
\begingroup
\small
\setlength{\tabcolsep}{5pt}
\renewcommand{\arraystretch}{1.12}
\begin{tabular*}{\columnwidth}{@{\extracolsep{\fill}}lccc@{}}
\toprule
LLM / Agent & Human 1 & Human 2 & Human 3 \\
\midrule
claude-fable-5   & 0.1522 & 0.5249 & 0.0421 \\
gpt-5.6-sol      & 0.2461 & 0.3658 & 0.1844 \\
qwen3.8-max      & 0.1835 & 0.4103 & 0.0888 \\
\midrule
ScholarPeer     & 0.1826 & 0.2198 & 0.1860 \\
AutoResearchClaw & 0.1345 & 0.2517 & 0.1471 \\
\bottomrule
\end{tabular*}
\endgroup
\end{table}

\section{Results of Ablation Study}
\label{sec:app_Results of Ablation Study}
We assess deliberation and weighting through R1--R2 comparisons (Table~\ref{tab:ablation_r1_r2}) and direct R2--R2 comparisons (Table~\ref{tab:ablation_r2_r2}). The full model is more often preferred on overall paper quality than the variant without deliberation across both backbones, and outperforms both ablations within PaperOrchestra. The benefits of weighting and effects on literature review quality vary across settings.

\begin{table*}[t]
\centering
\caption{
Comparison of R1 and R2 for the full model and its ablations.
Each cell reports R1 Win / Tie / R2 Win (\%).
Bold marks the highest R2 win rate within each backbone.
}
\label{tab:ablation_r1_r2}
\begingroup
\setlength{\tabcolsep}{2pt}
\renewcommand{\arraystretch}{1.3}
\scriptsize
\resizebox{\textwidth}{!}{%
\begin{tabular}{@{}ccc
  *{3}{c}
  >{\columncolor{overallorange}}c
  *{5}{c}
  >{\columncolor{overallorange}[\tabcolsep][0pt]}c@{}}
\toprule
\multirow[c]{2}{*}{Backbone}
& \multirow[c]{2}{*}{Judge}
& \multirow[c]{2}{*}{Comparison}
& \multicolumn{4}{c}{Literature Review Quality}
& \multicolumn{6}{c}{Paper Quality} \\
\cmidrule(lr){4-7}
\cmidrule(lr){8-13}
& &
& \shortstack[c]{Organization\\\& Writing}
& \shortstack[c]{Positioning\\\& Novelty}
& \shortstack[c]{Relevance\\\& Focus}
& \shortstack[c]{Overall Literature\\Review Quality}
& \shortstack[c]{Scientific\\Depth}
& \shortstack[c]{Technical\\Execution}
& \shortstack[c]{Logical\\Flow}
& \shortstack[c]{Clarity of\\Writing}
& \shortstack[c]{Presentation\\of Evidence}
& \shortstack[c]{Overall Paper\\Quality} \\
\midrule

\multirow[c]{9}{*}{ARIS}
& \multirow[c]{3}{*}{claude-fable-5}
& $w/o$ deliberation
& 5.0/65.0/30.0
& 0.0/45.0/55.0
& 2.5/75.0/22.5
& \cellcolor{overallorange}7.5/37.5/55.0
& 0.0/5.0/\textbf{95.0}
& 0.0/25.0/75.0
& 2.5/35.0/\textbf{62.5}
& 2.5/42.5/55.0
& 0.0/22.5/77.5
& \cellcolor{overallorange}2.5/5.0/92.5 \\

& & $w/o$ weighting
& 2.5/52.5/\textbf{45.0}
& 0.0/25.0/\textbf{75.0}
& 0.0/55.0/\textbf{45.0}
& \cellcolor{overallorange}0.0/27.5/\textbf{72.5}
& 0.0/10.0/90.0
& 0.0/10.0/\textbf{90.0}
& 2.5/35.0/\textbf{62.5}
& 2.5/37.5/60.0
& 0.0/7.5/\textbf{92.5}
& \cellcolor{overallorange}0.0/5.0/\textbf{95.0} \\

& & \cellcolor{oursblue}Ours
& \cellcolor{oursblue}0.0/90.0/10.0
& \cellcolor{oursblue}0.0/47.5/52.5
& \cellcolor{oursblue}2.5/72.5/25.0
& \cellcolor{overallorange}2.5/27.5/70.0
& \cellcolor{oursblue}0.0/45.0/55.0
& \cellcolor{oursblue}0.0/67.5/32.5
& \cellcolor{oursblue}0.0/77.5/22.5
& \cellcolor{oursblue}0.0/37.5/\textbf{62.5}
& \cellcolor{oursblue}0.0/32.5/67.5
& \cellcolor{overallorange}0.0/7.5/92.5 \\

\cmidrule(lr){2-13}

& \multirow[c]{3}{*}{gpt-5.6-sol}
& $w/o$ deliberation
& 12.5/40.0/47.5
& 2.5/47.5/50.0
& 2.5/72.5/25.0
& \cellcolor{overallorange}12.5/35.0/52.5
& 0.0/0.0/\textbf{100.0}
& 0.0/12.5/87.5
& 12.5/22.5/65.0
& 17.5/32.5/\textbf{50.0}
& 0.0/20.0/80.0
& \cellcolor{overallorange}2.5/7.5/90.0 \\

& & $w/o$ weighting
& 7.5/35.0/\textbf{57.5}
& 0.0/25.0/\textbf{75.0}
& 10.0/35.0/\textbf{55.0}
& \cellcolor{overallorange}7.5/25.0/\textbf{67.5}
& 0.0/0.0/\textbf{100.0}
& 0.0/0.0/\textbf{100.0}
& 10.0/10.0/\textbf{80.0}
& 15.0/35.0/\textbf{50.0}
& 0.0/0.0/\textbf{100.0}
& \cellcolor{overallorange}0.0/0.0/\textbf{100.0} \\

& & \cellcolor{oursblue}Ours
& \cellcolor{oursblue}12.5/75.0/12.5
& \cellcolor{oursblue}7.5/65.0/27.5
& \cellcolor{oursblue}5.0/85.0/10.0
& \cellcolor{overallorange}7.5/70.0/22.5
& \cellcolor{oursblue}0.0/0.0/\textbf{100.0}
& \cellcolor{oursblue}0.0/12.5/87.5
& \cellcolor{oursblue}12.5/37.5/50.0
& \cellcolor{oursblue}27.5/50.0/22.5
& \cellcolor{oursblue}0.0/10.0/90.0
& \cellcolor{overallorange}0.0/2.5/97.5 \\

\cmidrule(lr){2-13}

& \multirow[c]{3}{*}{qwen3.8-max}
& $w/o$ deliberation
& 7.5/42.5/50.0
& 0.0/42.5/57.5
& 2.5/62.5/35.0
& \cellcolor{overallorange}7.5/30.0/62.5
& 0.0/2.5/97.5
& 0.0/17.5/82.5
& 5.0/12.5/82.5
& 12.5/15.0/72.5
& 0.0/25.0/75.0
& \cellcolor{overallorange}2.5/2.5/95.0 \\

& & $w/o$ weighting
& 2.5/27.5/\textbf{70.0}
& 2.5/15.0/\textbf{82.5}
& 2.5/32.5/\textbf{65.0}
& \cellcolor{overallorange}2.5/10.0/\textbf{87.5}
& 0.0/0.0/\textbf{100.0}
& 0.0/0.0/\textbf{100.0}
& 2.5/12.5/\textbf{85.0}
& 5.0/10.0/\textbf{85.0}
& 0.0/2.5/\textbf{97.5}
& \cellcolor{overallorange}0.0/2.5/\textbf{97.5} \\

& & \cellcolor{oursblue}Ours
& \cellcolor{oursblue}15.0/62.5/22.5
& \cellcolor{oursblue}2.5/62.5/35.0
& \cellcolor{oursblue}10.0/75.0/15.0
& \cellcolor{overallorange}10.0/55.0/35.0
& \cellcolor{oursblue}0.0/5.0/95.0
& \cellcolor{oursblue}0.0/12.5/87.5
& \cellcolor{oursblue}7.5/20.0/72.5
& \cellcolor{oursblue}7.5/27.5/65.0
& \cellcolor{oursblue}0.0/12.5/87.5
& \cellcolor{overallorange}0.0/2.5/\textbf{97.5} \\

\midrule

\multirow[c]{9}{*}{PaperOrchestra}
& \multirow[c]{3}{*}{claude-fable-5}
& $w/o$ deliberation
& 0.0/85.0/15.0
& 2.5/75.0/22.5
& 0.0/95.0/5.0
& \cellcolor{overallorange}2.5/70.0/27.5
& 5.0/25.0/70.0
& 0.0/45.0/55.0
& 5.0/67.5/27.5
& 0.0/47.5/\textbf{52.5}
& 5.0/62.5/32.5
& \cellcolor{overallorange}5.0/12.5/82.5 \\

& & $w/o$ weighting
& 10.0/62.5/\textbf{27.5}
& 2.5/77.5/20.0
& 2.5/85.0/12.5
& \cellcolor{overallorange}10.0/57.5/32.5
& 15.0/47.5/37.5
& 5.0/75.0/20.0
& 10.0/55.0/35.0
& 2.5/62.5/35.0
& 5.0/72.5/22.5
& \cellcolor{overallorange}15.0/32.5/52.5 \\

& & \cellcolor{oursblue}Ours
& \cellcolor{oursblue}17.5/72.5/10.0
& \cellcolor{oursblue}5.0/55.0/\textbf{40.0}
& \cellcolor{oursblue}15.0/60.0/\textbf{25.0}
& \cellcolor{overallorange}17.5/40.0/\textbf{42.5}
& \cellcolor{oursblue}2.5/7.5/\textbf{90.0}
& \cellcolor{oursblue}2.5/22.5/\textbf{75.0}
& \cellcolor{oursblue}10.0/40.0/\textbf{50.0}
& \cellcolor{oursblue}17.5/52.5/30.0
& \cellcolor{oursblue}10.0/35.0/\textbf{55.0}
& \cellcolor{overallorange}5.0/7.5/\textbf{87.5} \\

\cmidrule(lr){2-13}

& \multirow[c]{3}{*}{gpt-5.6-sol}
& $w/o$ deliberation
& 10.0/75.0/15.0
& 15.0/57.5/27.5
& 0.0/77.5/22.5
& \cellcolor{overallorange}15.0/57.5/27.5
& 12.5/2.5/85.0
& 10.0/17.5/72.5
& 7.5/42.5/50.0
& 5.0/55.0/\textbf{40.0}
& 20.0/25.0/55.0
& \cellcolor{overallorange}12.5/2.5/85.0 \\

& & $w/o$ weighting
& 10.0/62.5/\textbf{27.5}
& 10.0/52.5/\textbf{37.5}
& 5.0/60.0/\textbf{35.0}
& \cellcolor{overallorange}12.5/37.5/\textbf{50.0}
& 20.0/35.0/45.0
& 10.0/52.5/37.5
& 12.5/47.5/40.0
& 5.0/60.0/35.0
& 27.5/27.5/45.0
& \cellcolor{overallorange}30.0/15.0/55.0 \\

& & \cellcolor{oursblue}Ours
& \cellcolor{oursblue}15.0/82.5/2.5
& \cellcolor{oursblue}5.0/60.0/35.0
& \cellcolor{oursblue}2.5/85.0/12.5
& \cellcolor{overallorange}10.0/52.5/37.5
& \cellcolor{oursblue}2.5/0.0/\textbf{97.5}
& \cellcolor{oursblue}0.0/15.0/\textbf{85.0}
& \cellcolor{oursblue}12.5/35.0/\textbf{52.5}
& \cellcolor{oursblue}32.5/40.0/27.5
& \cellcolor{oursblue}2.5/30.0/\textbf{67.5}
& \cellcolor{overallorange}2.5/10.0/\textbf{87.5} \\

\cmidrule(lr){2-13}

& \multirow[c]{3}{*}{qwen3.8-max}
& $w/o$ deliberation
& 7.5/62.5/30.0
& 15.0/55.0/30.0
& 7.5/65.0/27.5
& \cellcolor{overallorange}12.5/52.5/35.0
& 10.0/5.0/85.0
& 10.0/12.5/\textbf{77.5}
& 7.5/37.5/55.0
& 5.0/42.5/52.5
& 12.5/27.5/60.0
& \cellcolor{overallorange}12.5/5.0/82.5 \\

& & $w/o$ weighting
& 15.0/42.5/\textbf{42.5}
& 15.0/47.5/37.5
& 2.5/65.0/\textbf{32.5}
& \cellcolor{overallorange}17.5/35.0/47.5
& 17.5/37.5/45.0
& 7.5/55.0/37.5
& 12.5/42.5/45.0
& 10.0/37.5/52.5
& 10.0/47.5/42.5
& \cellcolor{overallorange}17.5/30.0/52.5 \\

& & \cellcolor{oursblue}Ours
& \cellcolor{oursblue}22.5/45.0/32.5
& \cellcolor{oursblue}15.0/32.5/\textbf{52.5}
& \cellcolor{oursblue}7.5/65.0/27.5
& \cellcolor{overallorange}17.5/30.0/\textbf{52.5}
& \cellcolor{oursblue}2.5/2.5/\textbf{95.0}
& \cellcolor{oursblue}2.5/30.0/67.5
& \cellcolor{oursblue}5.0/15.0/\textbf{80.0}
& \cellcolor{oursblue}10.0/25.0/65.0
& \cellcolor{oursblue}2.5/25.0/\textbf{72.5}
& \cellcolor{overallorange}5.0/5.0/\textbf{90.0} \\

\bottomrule
\end{tabular}%
}
\endgroup
\end{table*}


\begin{table*}[t]
\centering
\caption{
Comparison of R2 manuscripts revised by ablated models and ours.
Each cell reports Ablation Win / Tie / Ours Win (\%).
Bold marks the higher win rate between the ablation and ours.
Orange columns highlight overall metrics obtained by equally weighting
the subdimensions.
}
\label{tab:ablation_r2_r2}
\begingroup
\setlength{\tabcolsep}{2pt}
\renewcommand{\arraystretch}{1.15}
\scriptsize
\resizebox{\textwidth}{!}{%
\begin{tabular}{@{}ccc
  *{3}{c}
  >{\columncolor{overallorange}}c
  *{5}{c}
  >{\columncolor{overallorange}[\tabcolsep][0pt]}c@{}}
\toprule
\multirow[c]{2}{*}{Backbone}
& \multirow[c]{2}{*}{Judge}
& \multirow[c]{2}{*}{Comparison}
& \multicolumn{4}{c}{Literature Review Quality}
& \multicolumn{6}{c}{Paper Quality} \\
\cmidrule(lr){4-7}
\cmidrule(lr){8-13}
& &
& \shortstack[c]{Organization\\\& Writing}
& \shortstack[c]{Positioning\\\& Novelty}
& \shortstack[c]{Relevance\\\& Focus}
& \shortstack[c]{Overall Literature\\Review Quality}
& \shortstack[c]{Scientific\\Depth}
& \shortstack[c]{Technical\\Execution}
& \shortstack[c]{Logical\\Flow}
& \shortstack[c]{Clarity of\\Writing}
& \shortstack[c]{Presentation\\of Evidence}
& \shortstack[c]{Overall Paper\\Quality} \\
\midrule

\multirow[c]{6}{*}{ARIS}
& \multirow[c]{2}{*}{claude-fable-5}
& $w/o$ deliberation vs. Ours
& \textbf{27.5}/62.5/10.0
& \textbf{27.5}/52.5/20.0
& \textbf{12.5}/77.5/10.0
& \cellcolor{overallorange}\textbf{35.0}/37.5/27.5
& 20.0/37.5/\textbf{42.5}
& 15.0/57.5/\textbf{27.5}
& \textbf{22.5}/60.0/17.5
& \textbf{22.5}/72.5/5.0
& 25.0/40.0/\textbf{35.0}
& \cellcolor{overallorange}30.0/25.0/\textbf{45.0} \\

& & $w/o$ weighting vs. Ours
& \textbf{30.0}/65.0/5.0
& \textbf{42.5}/52.5/5.0
& \textbf{27.5}/70.0/2.5
& \cellcolor{overallorange}\textbf{45.0}/47.5/7.5
& \textbf{60.0}/32.5/7.5
& \textbf{55.0}/40.0/5.0
& \textbf{30.0}/55.0/15.0
& \textbf{30.0}/60.0/10.0
& \textbf{50.0}/40.0/10.0
& \cellcolor{overallorange}\textbf{65.0}/25.0/10.0 \\

\cmidrule(lr){2-13}

& \multirow[c]{2}{*}{gpt-5.6-sol}
& $w/o$ deliberation vs. Ours
& \textbf{42.5}/45.0/12.5
& \textbf{40.0}/45.0/15.0
& \textbf{25.0}/65.0/10.0
& \cellcolor{overallorange}\textbf{47.5}/30.0/22.5
& 30.0/35.0/\textbf{35.0}
& 32.5/32.5/\textbf{35.0}
& \textbf{47.5}/27.5/25.0
& \textbf{37.5}/42.5/20.0
& \textbf{37.5}/27.5/35.0
& \cellcolor{overallorange}40.0/12.5/\textbf{47.5} \\

& & $w/o$ weighting vs. Ours
& \textbf{42.5}/50.0/7.5
& \textbf{57.5}/37.5/5.0
& \textbf{35.0}/60.0/5.0
& \cellcolor{overallorange}\textbf{65.0}/27.5/7.5
& \textbf{72.5}/25.0/2.5
& \textbf{70.0}/25.0/5.0
& \textbf{37.5}/42.5/20.0
& \textbf{30.0}/47.5/22.5
& \textbf{70.0}/15.0/15.0
& \cellcolor{overallorange}\textbf{75.0}/7.5/17.5 \\

\cmidrule(lr){2-13}

& \multirow[c]{2}{*}{qwen3.8-max}
& $w/o$ deliberation vs. Ours
& \textbf{30.0}/57.5/12.5
& \textbf{32.5}/55.0/12.5
& \textbf{22.5}/67.5/10.0
& \cellcolor{overallorange}\textbf{35.0}/47.5/17.5
& 25.0/45.0/\textbf{30.0}
& 15.0/40.0/\textbf{45.0}
& \textbf{42.5}/40.0/17.5
& \textbf{42.5}/32.5/25.0
& 17.5/45.0/\textbf{37.5}
& \cellcolor{overallorange}30.0/22.5/\textbf{47.5} \\

& & $w/o$ weighting vs. Ours
& \textbf{42.5}/52.5/5.0
& \textbf{45.0}/45.0/10.0
& \textbf{37.5}/52.5/10.0
& \cellcolor{overallorange}\textbf{42.5}/47.5/10.0
& \textbf{62.5}/32.5/5.0
& \textbf{62.5}/30.0/7.5
& \textbf{32.5}/52.5/15.0
& \textbf{40.0}/37.5/22.5
& \textbf{55.0}/32.5/12.5
& \cellcolor{overallorange}\textbf{70.0}/17.5/12.5 \\

\midrule

\multirow[c]{6}{*}{PaperOrchestra}
& \multirow[c]{2}{*}{claude-fable-5}
& $w/o$ deliberation vs. Ours
& \textbf{27.5}/55.0/17.5
& 17.5/42.5/\textbf{40.0}
& \textbf{17.5}/70.0/12.5
& \cellcolor{overallorange}32.5/32.5/\textbf{35.0}
& 20.0/27.5/\textbf{52.5}
& 30.0/40.0/\textbf{30.0}
& 32.5/35.0/\textbf{32.5}
& \textbf{40.0}/37.5/22.5
& 27.5/42.5/\textbf{30.0}
& \cellcolor{overallorange}37.5/15.0/\textbf{47.5} \\

& & $w/o$ weighting vs. Ours
& \textbf{30.0}/60.0/10.0
& \textbf{35.0}/40.0/25.0
& \textbf{32.5}/52.5/15.0
& \cellcolor{overallorange}\textbf{45.0}/32.5/22.5
& 22.5/12.5/\textbf{65.0}
& 15.0/35.0/\textbf{50.0}
& 35.0/30.0/\textbf{35.0}
& \textbf{37.5}/47.5/15.0
& 32.5/22.5/\textbf{45.0}
& \cellcolor{overallorange}30.0/10.0/\textbf{60.0} \\

\cmidrule(lr){2-13}

& \multirow[c]{2}{*}{gpt-5.6-sol}
& $w/o$ deliberation vs. Ours
& \textbf{35.0}/45.0/20.0
& 27.5/32.5/\textbf{40.0}
& \textbf{30.0}/47.5/22.5
& \cellcolor{overallorange}42.5/15.0/\textbf{42.5}
& 20.0/17.5/\textbf{62.5}
& 40.0/17.5/\textbf{42.5}
& 32.5/35.0/\textbf{32.5}
& 30.0/40.0/\textbf{30.0}
& 20.0/37.5/\textbf{42.5}
& \cellcolor{overallorange}30.0/15.0/\textbf{55.0} \\

& & $w/o$ weighting vs. Ours
& \textbf{45.0}/52.5/2.5
& \textbf{37.5}/27.5/35.0
& \textbf{32.5}/50.0/17.5
& \cellcolor{overallorange}\textbf{47.5}/27.5/25.0
& 20.0/12.5/\textbf{67.5}
& 17.5/25.0/\textbf{57.5}
& 30.0/35.0/\textbf{35.0}
& \textbf{37.5}/35.0/27.5
& 25.0/17.5/\textbf{57.5}
& \cellcolor{overallorange}27.5/5.0/\textbf{67.5} \\

\cmidrule(lr){2-13}

& \multirow[c]{2}{*}{qwen3.8-max}
& $w/o$ deliberation vs. Ours
& \textbf{32.5}/50.0/17.5
& 25.0/30.0/\textbf{45.0}
& 22.5/50.0/\textbf{27.5}
& \cellcolor{overallorange}35.0/20.0/\textbf{45.0}
& 30.0/15.0/\textbf{55.0}
& \textbf{37.5}/30.0/32.5
& 27.5/32.5/\textbf{40.0}
& 32.5/35.0/\textbf{32.5}
& 22.5/37.5/\textbf{40.0}
& \cellcolor{overallorange}40.0/10.0/\textbf{50.0} \\

& & $w/o$ weighting vs. Ours
& \textbf{35.0}/45.0/20.0
& 30.0/35.0/\textbf{35.0}
& \textbf{27.5}/47.5/25.0
& \cellcolor{overallorange}37.5/22.5/\textbf{40.0}
& 25.0/12.5/\textbf{62.5}
& 30.0/12.5/\textbf{57.5}
& 30.0/20.0/\textbf{50.0}
& 32.5/32.5/\textbf{35.0}
& 35.0/20.0/\textbf{45.0}
& \cellcolor{overallorange}35.0/7.5/\textbf{57.5} \\

\bottomrule
\end{tabular}%
}
\endgroup
\end{table*}

\section{Case Study}
\label{sec:case-study}
This case study highlights the advantage of RoundtableReview in translating targeted feedback into concrete, reproducible revisions. AI Scientist-v2 requests general technical clarification, but its revision introduces a similarity threshold and top-$k$ sampling, altering the original nearest-neighbor procedure. GLM-5.2 suggests broader noise-modeling extensions, yet the corresponding passage remains unchanged. In contrast, RoundtableReview identifies specific reproducibility gaps and elicits a revision specifying the BERT encoder and representation, cosine similarity, an explicit label-selection equation, corruption probability, and binary-answer handling, with a reference to complete pseudocode. The revision also clarifies that no similarity threshold is used, preserving the original nearest-neighbor formulation. These improvements illustrate how RoundtableReview produces more actionable feedback and clearer methodological specifications while maintaining fidelity to the stated method.

\begin{table*}[p]
\centering
\caption{Case Study of comparison of excerpts from Our Method, AI-scientist-v2, and GLM-5.2 Agent Reviews: Original Paragraph, Review Comments, and Corresponding Revised Paragraph. Italicized text indicates changes relative to the original paragraph based on review comments.}
\label{tab:semantic-noise-case-study}
\begingroup
\fontsize{8.5}{10}\selectfont
\setlength{\tabcolsep}{4pt}
\renewcommand{\arraystretch}{1.08}
\emergencystretch=1em
\begin{adjustbox}{max totalsize={\textwidth}{0.82\textheight},keepaspectratio}
\begin{tabular}{@{}p{22mm}p{\dimexpr\textwidth-22mm-2\tabcolsep\relax}@{}}
\hline
\noalign{\vskip 4pt}
\textbf{Original} &
\textbf{Semantic Noise Simulation}\par\smallskip
Standard noise simulation uses random label shuffling, which produces medically implausible mislabeling (e.g., ``X-ray'' for a pathology question). We propose a more realistic simulation: for each answer $a_k$, we encode all possible answers using a pre-trained BERT model and replace $a_k$ with its nearest neighbor in BERT embedding space, creating a semantically plausible but incorrect label. Figure 3 illustrates that semantic noise pairs retain feature-space proximity, unlike random noise.\par
\vspace{4pt}
\hrule width 0.20\linewidth height 0.4pt
\vspace{4pt}
\noindent
\textbf{Noise protocols}\par\smallskip
We evaluate under three noise conditions: 10\% Semantic Noise, 20\% Semantic Noise (via our BERT-proximity simulation), and 20\% Random Noise (uniform label shuffling). Test sets remain clean in all settings. The \textbf{Upper Bound} is performance with clean training labels (VQA-RAD: 79.13\%; PathVQA: 72.61\% overall).
\\
\noalign{\vskip 4pt}
\hline
\noalign{\vskip 4pt}
\textbf{AI-scientist-v2} &
\textbf{Review Comment}\par\smallskip
The paper lacks clarity in explaining certain technical details, such as the architecture and hyperparameter choices for the Noisy Label Refinement (NLR) module and the setup of the semantic noise simulation protocol.
\par
\vspace{4pt}
\hrule width \linewidth height 0.4pt
\vspace{4pt}
\noindent
\textbf{Revised Paragraph}\par\smallskip
\textbf{Semantic Noise Simulation}\quad Unchanged from the original.\par
\vspace{4pt}
\hrule width 0.20\linewidth height 0.4pt
\vspace{4pt}
\noindent
\textbf{Noise protocols}\par\smallskip
We evaluate under three noise conditions: 10\% Semantic Noise, 20\% Semantic Noise (via our BERT-proximity simulation), and 20\% Random Noise (uniform label shuffling). \textit{The semantic noise simulation protocol constructs corrupted labels by replacing the ground-truth answer with its nearest neighbor in BERT embedding space, subject to the constraint that the replacement is semantically plausible (cosine similarity $>0.6$) but distinct from the correct answer. Concretely, for each training sample selected for corruption, we encode all candidate answers with a frozen BERT encoder, retrieve the top-$k$ ($k=5$) nearest neighbors of the ground-truth answer embedding, and sample the replacement uniformly from this set. This reflects realistic annotation errors where a clinician's incorrect label is at least semantically adjacent to the correct one (e.g., ``left'' vs. ``right'', ``high magnification'' vs. ``low magnification'').} Test sets remain clean in all settings. The \textbf{Upper Bound} is performance with clean training labels (VQA-RAD: 79.13\%; PathVQA: 72.61\% overall).
\\
\noalign{\vskip 4pt}
\hline
\noalign{\vskip 4pt}
\textbf{GLM-5.2} &
\textbf{Review Comment}\par\smallskip
The semantic noise simulation protocol, while a valuable contribution, is limited to nearest-neighbor replacement in BERT space. Real medical annotation errors may involve hierarchical relationships (e.g., `lobar pneumonia' vs `bronchopneumonia') or multi-attribute confusion. The protocol could be enriched with multiple noise levels or structured noise patterns.
\par
\vspace{4pt}
\hrule width \linewidth height 0.4pt
\vspace{4pt}
\noindent
\textbf{Revised Paragraph}\par\smallskip
\textbf{Semantic Noise Simulation.}\quad Unchanged from the original.\par
\vspace{4pt}
\hrule width 0.20\linewidth height 0.4pt
\vspace{4pt}
\noindent
\textbf{Noise protocols.}\quad Unchanged from the original.
\\
\noalign{\vskip 4pt}
\hline
\noalign{\vskip 4pt}
\textbf{Ours} &
\textbf{Review Comment}\par\smallskip
The method description lacks critical algorithmic details—including the diffusion noise schedule, exact Robust Focal Loss formulation, EMA blending formula, and the semantic noise simulation protocol's BERT variant, similarity metric, and threshold parameters—preventing independent reproduction; the authors should provide complete mathematical definitions, step-by-step pseudocode, and hyperparameter sensitivity ranges in supplementary material to ensure graduate-level reproducibility.
\par
\vspace{4pt}
\hrule width \linewidth height 0.4pt
\vspace{4pt}
\noindent
\textbf{Revised Paragraph}\par\smallskip
\textbf{Semantic Noise Simulation.}\par\smallskip
Standard noise simulation uses random label shuffling, which produces medically implausible mislabeling (e.g., ``X-ray'' for a pathology question). We propose a more realistic simulation: for each answer $a_k$, we encode all possible answers using \textit{BERT-Base ([CLS] token of the final layer, $\ifdefined\mathbb\mathbb{R}\else\mathbf{R}\fi^{768}$)} and replace $a_k$ with its nearest neighbor in \textit{cosine similarity space. Formally, the noisy label is $a_k^*=\mathop{\mathrm{arg\,max}}\limits_{a\ne a_k,\,a\in\mathcal{A}}\cos(e_{a_k},e_a)$, where the nearest neighbor is taken as a hard selection (no similarity threshold; see Sec. A.1 for complete pseudocode). With probability $\eta$ (noise ratio), we replace $a_k$ with $a_k^*$; otherwise $a_k$ is kept. For closed-end Yes/No questions, symmetric binary noise is used since the two-class vocabulary does not admit meaningful semantic nearest-neighbor computation.} Figure 3 illustrates that semantic noise pairs retain feature-space proximity, unlike random noise.\par
\vspace{4pt}
\hrule width 0.20\linewidth height 0.4pt
\vspace{4pt}
\noindent
\textbf{Noise protocols.}\quad Unchanged from the original.
\\
\noalign{\vskip 4pt}
\hline
\end{tabular}
\end{adjustbox}
\endgroup
\end{table*}

\end{document}